\documentclass[11pt]{article}
\usepackage[preprint]{acl}
\usepackage{times}
\usepackage{latexsym}
\usepackage[T1]{fontenc}
\usepackage[utf8]{inputenc}
\usepackage{microtype}
\usepackage{framed}
\usepackage{mdframed}
\usepackage{listings}
\usepackage{verbatim}
\usepackage{inconsolata}
\usepackage{tikz}

\usepackage{microtype}
\usepackage{graphicx}
\usepackage{subfigure}
\usepackage{booktabs} 
\usepackage{algorithm}
\usepackage{algorithmic}
\usepackage{csquotes}

\usepackage{hyperref}

\usepackage{amsmath}
\usepackage{amssymb}
\usepackage{mathtools}
\usepackage{amsthm}

\usepackage[capitalize,noabbrev]{cleveref}

\usepackage{xcolor}
\usepackage{listings}
\usepackage{graphicx}
\usepackage[most]{tcolorbox}
\usepackage{lipsum}
\usepackage{multicol}

\theoremstyle{plain}

\theoremstyle{definition}

\theoremstyle{remark}

\usepackage[textsize=tiny]{todonotes}
\usepackage{enumitem}
\usepackage{booktabs}
\usepackage{pifont}
\usepackage{makecell}

\title{Investigating Pedagogical Teacher and Student LLM Agents: Genetic Adaptation Meets Retrieval‑Augmented Generation Across Learning Styles}

\author{Debdeep Sanyal\textsuperscript{1}, Agniva Maiti\textsuperscript{1}, Umakanta Maharana\textsuperscript{1}, Dhruv Kumar\textsuperscript{2}, \\
\textbf{Ankur Mali\textsuperscript{3}, C. Lee Giles\textsuperscript{4}, Murari Mandal\textsuperscript{1}} \\
\textsuperscript{1} RespAI Lab, KIIT Bhubaneswar, \\
\textsuperscript{2} BITS Pilani, \textsuperscript{3} University of South Florida, \textsuperscript{4} Pennsylvania State University\\
\textbf{Correspondence:} \texttt{murari.mandalfcs@kiit.ac.in}
}

\begin{document}
\maketitle

\begin{abstract}
Effective teaching requires adapting instructional strategies to accommodate the diverse cognitive and behavioral profiles of students, a persistent challenge in education and teacher training. While Large Language Models (LLMs) offer promise as tools to simulate such complex pedagogical environments, current simulation frameworks are limited in two key respects: (1) they often reduce students to static knowledge profiles, and (2) they lack adaptive mechanisms for modeling teachers who evolve their strategies in response to student feedback.  To address these gaps, \textbf{we introduce a novel simulation framework that integrates LLM-based heterogeneous student agents with a self-optimizing teacher agent}. The teacher agent’s pedagogical policy is dynamically evolved using a genetic algorithm, allowing it to discover and refine effective teaching strategies based on the aggregate performance of diverse learners.  In addition, \textbf{we propose Persona-RAG}, a Retrieval-Augmented Generation module that enables student agents to retrieve knowledge tailored to their individual learning styles. Persona-RAG preserves the retrieval accuracy of standard RAG baselines while enhancing personalization, an essential factor in modeling realistic educational scenarios. Through extensive experiments, we demonstrate how our framework supports the emergence of distinct and interpretable teaching patterns when interacting with varied student populations. Our results highlight the potential of LLM-driven simulations to inform adaptive teaching practices and provide a testbed for training human educators in controlled, data-driven environments.
\end{abstract}
\begin{figure*}[t]
     \centering
     \includegraphics[width=0.8\textwidth]{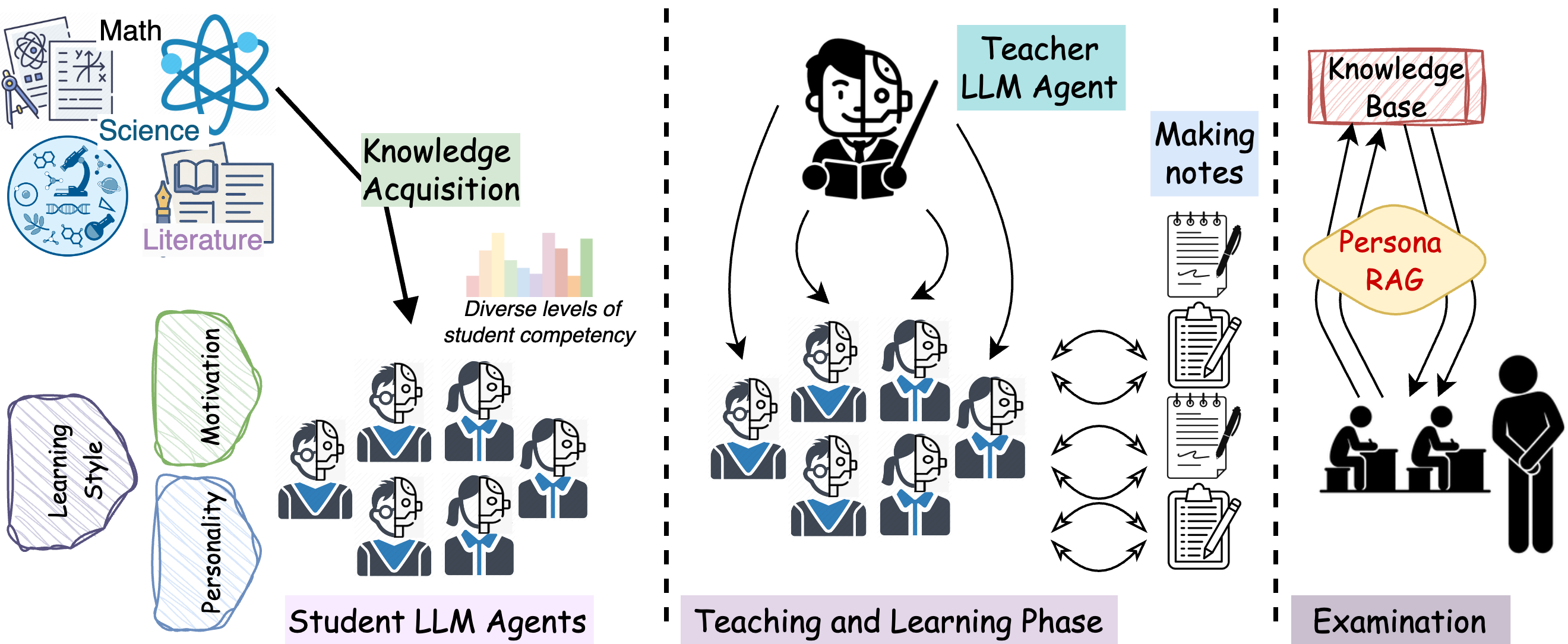}
     \caption{Complete pipeline for our pedagogical setup. Initially, the student agents are prepared with individual knowledge bases, containing prerequisite knowledge for Math, Science and English as per their subject aptitudes. The teacher agent teaches a topic from these three subjects building upon the prerequisite knowledge of the students, and the students are assessed on how well they have learned the topics. The teacher agent optimizes to increase the average score of the classroom.}
     \label{fig:pipeline}
     \vspace{-1\baselineskip}
\end{figure*}

\section{Introduction}
\label{sec:intro}
Effective education hinges on a teacher's ability to adapt their methods to the diverse needs of students, accounting for variations in aptitude, learning styles, and personality \cite{Agustrianita2019TeachersPO,Hasib2021LearnerAIA,Keshavarz2019TheEOA,Liu2025OneSD}. Mastering this adaptive pedagogy remains a key challenge in teacher training, requiring deep insight into student-teacher dynamics and individualized learning processes.

Recent advances in large language models (LLMs) \cite{qwen2025qwen25technicalreport,sanyal2025alu,deepseekai2025deepseekv3technicalreport,gemmateam2025gemma3technicalreport,openai2024gpt4ocard,abdin2024phi4technicalreport} have enabled the construction of multi-agent educational simulations~\cite{Zhang2024SimulatingCE,Aperstein2025GenerativeAP,Yu2024FromMT,Chu2025LLMAF}, providing controlled, reproducible environments for studying and discovering optimal teaching strategies. These LLM-driven simulations offer tremendous potential to inform teacher training programs and personalize curricula based on how different students learn and respond to instruction \cite{Nwana1990,Liu_nwana}.

However, building effective simulations for adaptive pedagogy is non-trivial. Prior work has primarily focused on surface level interaction fidelity \cite{Zhang2024SimulatingCE,Aperstein2025GenerativeAP,Chu2025LLMAF}, while neglecting the crucial feedback loop where student learning outcomes actively guide pedagogical adjustments. Capturing this loop is technically challenging, given the high dimensional nature of lecture delivery, note-taking, assessment, and evaluation, all of which make gradient-based optimization infeasible. Furthermore, while Retrieval-Augmented Generation (RAG) \cite{lewis2021retrievalaugmentedgenerationknowledgeintensivenlp} enables LLMs to access external knowledge, existing RAG variants \cite{chan2024rqraglearningrefinequeries,gao2022precisezeroshotdenseretrieval,izacard2022atlasfewshotlearningretrieval} lack personalization mechanisms essential for modeling individual student retrieval behavior.

To address these limitations, we propose a simulation framework that integrates diverse LLM-based student agents and a self-evolving teacher agent. Our student model captures cognitive and behavioral diversity, incorporating six learning styles based on \textit{VARK} and \textit{Felder-Silverman} models: \textit{Read/Write, Visual, Auditory, Kinesthetic, Intuitive, Sequential/Analytical} \cite{Prithishkumar2014UnderstandingYS,felder} and five personality traits: \textit{Social, Diligent, Independent, Anxious, Curious} derived from a survey of university instructors. This modeling enables nuanced simulation of individual knowledge acquisition.

To evolve teaching strategies, we use genetic algorithms (GAs) \cite{alam2020geneticalgorithmreviewsimplementations,lee2025evolvingdeeperllmthinking}, which bypass gradient dependence by encoding pedagogical parameters (e.g., style, tone, structure) as chromosomes. We initialize a population of 500 teacher agents, each delivering lectures on randomized topics in \textit{Math}, \textit{Science}, and \textit{English}, and evolve them over 50 generations based on student learning performance.

During simulation, students take personalized notes reflecting their style and personality, and populate their knowledge bases accordingly. They are then tested using RAG-based open book assessments, with performance scores feeding back into the GA's fitness function. To improve information access, we introduce \textbf{Persona-RAG}, a novel RAG architecture inspired by cognitive theories of human problem-solving \cite{cognition1, cognition2, cognition3, cognition4, cognition5}. Persona-RAG first generates an intermediate problem solving strategy (a “plan”) before querying the knowledge base, tailoring retrieval to the student’s preferred reasoning path. As shown in Section~\ref{sec:exp}, this significantly improves answer accuracy on non recall questions.

We conduct comprehensive empirical validation. While no direct baselines exist for optimizing teacher strategies in such diverse multi-agent simulations, we demonstrate that the GA learns interpretable teaching behaviors that significantly improve class level performance. We further analyze adaptation patterns in homogeneous classrooms, showing how teacher strategies evolve based on shared student traits.

For retrieval, we compare Persona-RAG against strong RAG baselines, including Query Translation, Query Decomposition \cite{chan2024rqraglearningrefinequeries}, and HyDE \cite{gao2022precisezeroshotdenseretrieval}, and show improved performance on pedagogical tasks requiring multistep reasoning. To bridge simulation and real-world validation, we collect human feedback on generated lectures (\S\ref{sec:humaneval}), demonstrating that strategies learned in simulation are perceived as effective and pedagogically engaging.

\vspace{1mm}
\noindent\textbf{Key Contributions.}\\
\ding{182} \textbf{A self evolving teacher agent} optimized using genetic algorithms based on aggregate simulated student performance.\\
\ding{183} \textbf{A rich simulation of student diversity}, combining six learning styles and five personality types grounded in educational theory and faculty surveys.\\
\ding{184} \textbf{Persona-RAG}, a cognitively inspired retrieval method tailored to student specific reasoning processes in learning contexts.\\
\ding{185} \textbf{Extensive empirical validation}, showing measurable improvements in simulated learning outcomes and strong human perceived pedagogical value.

Our framework paves the way for scalable, personalized, and interpretable teacher training using AI based educational simulation.

\begin{figure*}[t]
     \centering
     \includegraphics[width=0.95\textwidth]{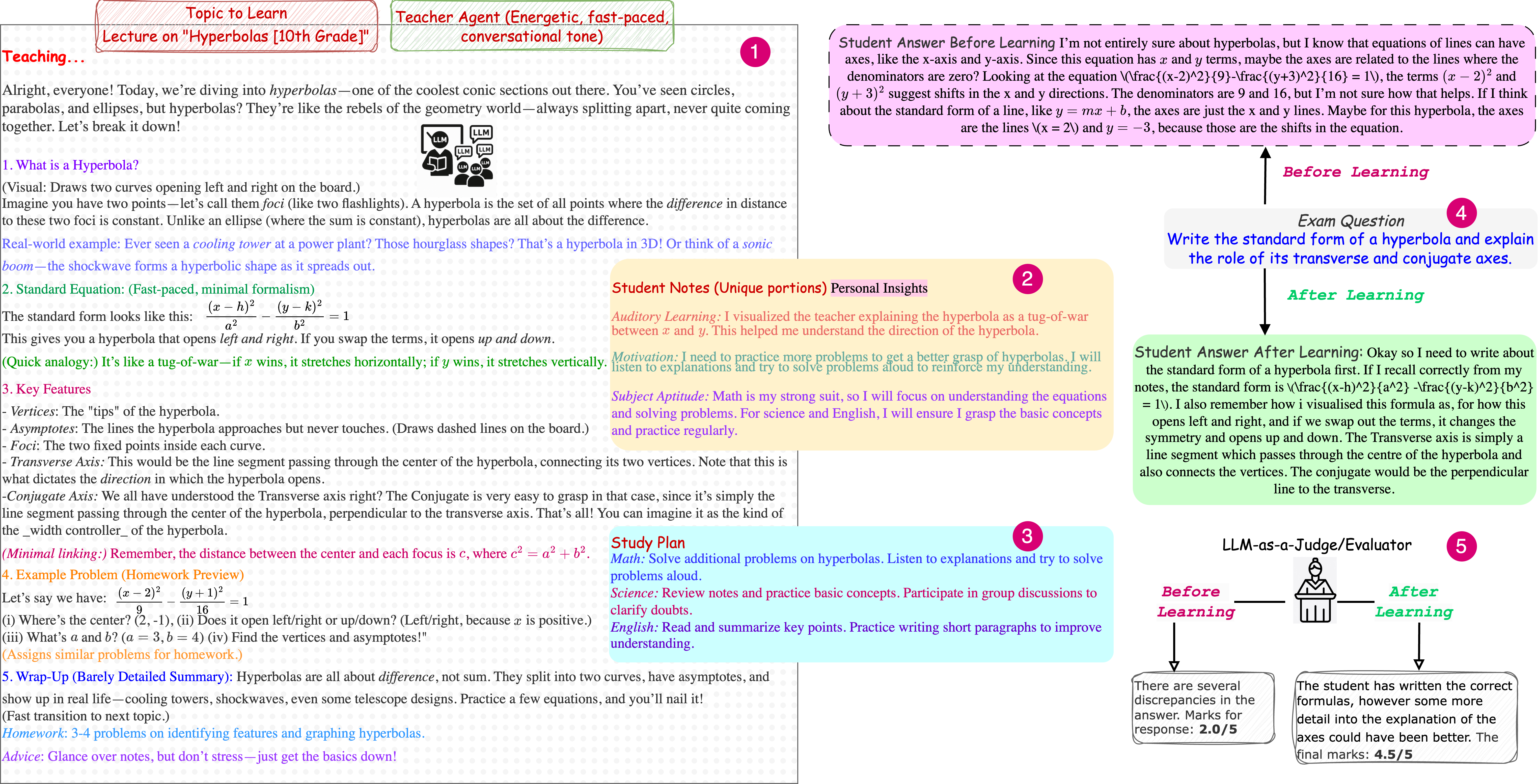}
     \caption{A complete example of a lecture from the teacher agent, the notes taken by the student agent, the student response to the assessment and the performance as scored by LLM-as-a-Judge. We observe that the teacher agent has learned to include real-world examples right after the first definition, provides analogies after presenting formulas for better intuitions, example problems and summaries at the end of the lecture for an overall recap. This leads to a complete and well rounded lecture. The student is able to answer effectively in the exam with the visualizations provided in the class.}
     \label{fig:sample_results}
     \vspace{-1\baselineskip}
\end{figure*}

\section{Related Work}
\paragraph{LLM-Based Multi-Agent Simulations.}  
Recent works have explored the use of LLM-based agents to simulate educational settings and learner personas~\cite{related1_1,related1_2,related1_3,related1_4,related1_5,related1_6}. Frameworks such as \textbf{SimClass}~\cite{related1_1} and \textbf{CGMI}~\cite{cgmi} construct richly populated classroom environments, emphasizing interaction fidelity and realism. However, these systems are primarily designed to simulate classroom dynamics, not to evaluate or optimize teaching strategies. Crucially, they lack an adaptive feedback loop that ties pedagogical actions to student learning outcomes. In contrast, our work integrates this missing loop: we use simulation not as an end, but as a controlled environment to discover and refine adaptive teaching strategies through performance driven optimization.

\paragraph{Simulation for Teacher Training.}  
Other lines of work, including \textbf{Generative Agent Design for Teacher Training (GAD-TT)}~\cite{gadtt} and \textbf{TutorUp}~\cite{Pan2025TutorUpWIA}, leverage LLM-based student agents to train human educators. These systems simulate realistic student behaviors to help novice teachers practice classroom management and instructional delivery. While valuable for human-in-the-loop pedagogy, these frameworks focus on training humans, not on optimizing the pedagogical behavior of the AI teacher itself. Our approach differs in that it treats the AI teacher as a learning agent, optimizing its strategy over generations based on simulated student performance.

\paragraph{Personalization and Learning Resource Optimization.}  
Systems like \textbf{GenMentor}~\cite{genmentor} personalize learning pathways by sequencing instructional content based on learner profiles. Similarly, \textbf{EduPlanner}~\cite{eduplanner} uses adversarial techniques to optimize lesson plans. These approaches optimize static artifacts (e.g., content sequencing), not live pedagogical behavior. Our framework instead focuses on the real-time, dynamic adaptation of a teacher agent’s delivery style, allowing it to tailor its behavior in response to diverse and evolving student states.
\paragraph{Positioning.}
In contrast to prior work that either simulates classroom dynamics or supports teacher training, our framework unifies both goals through a closed-loop system: it models cognitive and behavioral student diversity, and optimizes a dynamic teacher agent based on measured learning outcomes. This integration allows us to study adaptive pedagogy not just as a fixed process, but as an evolving strategy grounded in multi-agent interaction and performance feedback.

\begin{algorithm}[t]
\caption{\textsc{Persona-RAG}: Personalized Multi-Step Retrieval Guided by Student Characteristics}
\label{alg:Persona-RAG}
\begin{algorithmic}[1]

\REQUIRE 
\begin{itemize}[leftmargin=*,noitemsep,topsep=0pt]
    \item[] $Q \in \mathbb{S}$: Input question in natural language.
    \item[] $KB = \{d_1, \dots, d_n\}$: Student Knowledge Base, where $d_j \in \mathbb{S}$.
    \item[] $SC$: Student Characteristics (e.g., learning style, memory preference, abstraction level).
\end{itemize}

\ENSURE 
\begin{itemize}[leftmargin=*,noitemsep,topsep=0pt]
    \item[] $D_{\text{retrieved}}$: Personalized set of document chunks relevant to $Q$.
\end{itemize}

\vspace{1mm}
\textbf{Definitions:}
\begin{itemize}[leftmargin=*,noitemsep,topsep=0pt]
    \item $\textsc{Approach}(Q, SC) \rightarrow P = [p_1, \dots, p_m]$: Generates an individualized reasoning plan using $SC$.
    \item $\textsc{Retrieve}(p_i, KB) \rightarrow D_i \subseteq KB$: Retrieves documents from $KB$ semantically aligned with plan step $p_i$.
\end{itemize}

\vspace{1mm}
\STATE \textcolor{blue!70!black}{\textbf{Initialize:}} $D_{\text{retrieved}} \gets \emptyset$

\vspace{1mm}
\STATE \textcolor{blue!70!black}{\textbf{Step 1: Personalized Reasoning Plan:}}
\STATE $P \gets \textsc{Approach}(Q, SC)$ \COMMENT{Generates $m$ reasoning steps tailored to $SC$}

\vspace{1mm}
\STATE \textcolor{blue!70!black}{\textbf{Step 2: Multi-Step Retrieval:}}
\FOR{$i = 1$ to $m$}
    \STATE $D_i \gets \textsc{Retrieve}(p_i, KB)$
    \STATE $D_{\text{retrieved}} \gets D_{\text{retrieved}} \cup D_i$
\ENDFOR

\vspace{1mm}
\STATE \textcolor{blue!70!black}{\textbf{Return}} $D_{\text{retrieved}}$

\end{algorithmic}
\end{algorithm}

\section{Modeling Framework: Genetic Adaptation and RAG for Pedagogical LLM Agents}
\label{sec:methodology}
\subsection{Student Knowledge Base Construction}
\label{sec:kb_construction}
Creating realistic LLM-agent learning environments requires defining a controlled knowledge prerequisite, critical for modeling learning and evaluating teaching given LLMs' vast, undifferentiated pre-training knowledge. We establish a baseline for our simulated curriculum to accurately model the learning process and evaluate the effectiveness of different teaching approaches. Fundamental pre-undergraduate topics from English, Math, and Science, based on SAT, JEE, and Gaokao syllabi are selected for analysis (details in Appendix). The knowledge base (KB) comprises 20 topics in Mathematics (primarily Algebra and Geometry), 21 topics in Science (covering Physics, Chemistry, and Biology), and 37 topics related to English Literature and Grammar. Crucially, to simulate variable subject aptitudes among students which is a key element of student diversity, knowledge for each topic is structured across three distinct detail levels: \textbf{Level 1}: \textit{Provides a high-level overview and fundamental concepts, offering a basic understanding of the topic without delving into complex nuances.} \textbf{Level 2}: \textit{Includes more detailed descriptions of core foundations, essential principles, and introductory insights into slightly more advanced aspects.} \textbf{Level 3}: \textit{Consists of in-depth explanations, detailed analogies, derivations where applicable, and a thorough exploration of the topic's various components.} This tiered design allows instantiating student agents with differing initial knowledge per specific subject (e.g., Level 1 KB for low-aptitude Math topics), moving beyond impractical uniform aptitude. Providing variable understanding across a common set of topics fundamentally builds the diverse student population required for our adaptive teacher agent framework.

\begin{table*}[t]
    \centering
    \small
    \caption{Retrieval Accuracies on different types of questions compared against various RAG methods. We observe that all RAG methods perform well on simple recall based questions, however the accuracy drops for more complex ones. We observe that creative questions are especially more challenging since that requires nuanced retrieval and is highly dependent on personal styles. Persona-RAG has notably less variance than HyDE at similar accuracies. Others like Query Decomposition and Query Translation are good for for some question types but does not generalize well for others.}
    \label{tab:q_type}
    \begin{tabular}{l|ccccc}
        \toprule
        \textbf{Question Type} & \textbf{Vanilla RAG} & \textbf{Query Decomposition} & \textbf{Query Translation} & \textbf{HyDE} & \textbf{Persona-RAG}\\
        \midrule
        Simple Recall & $0.93 \pm 0.05$ & $0.91 \pm 0.03$ & $0.95 \pm 0.07$ & $0.92 \pm 0.09$ & $0.93 \pm 0.02$\\
        Conceptual & $0.74 \pm 0.08$ & $0.82 \pm 0.06$ & $0.77 \pm 0.04$ & $0.85 \pm 0.11$ & $0.88 \pm 0.07$\\
        Application Based & $0.79 \pm 0.02$ & $0.83 \pm 0.09$ & $0.86 \pm 0.05$ & $0.84 \pm 0.03$ & $0.81 \pm 0.03$\\
        Analysis Based & $0.68 \pm 0.10$ & $0.71 \pm 0.07$ & $0.69 \pm 0.02$ & $0.76 \pm 0.06$ & $0.85 \pm 0.04$\\
        Creative & $0.66 \pm 0.04$ & $0.69 \pm 0.11$ & $0.73 \pm 0.08$ & $0.77 \pm 0.09$ & $0.80 \pm 0.05$\\
        \bottomrule
    \end{tabular}
\end{table*}
\begin{figure}[t]
     \centering
     \includegraphics[width=0.40\textwidth]{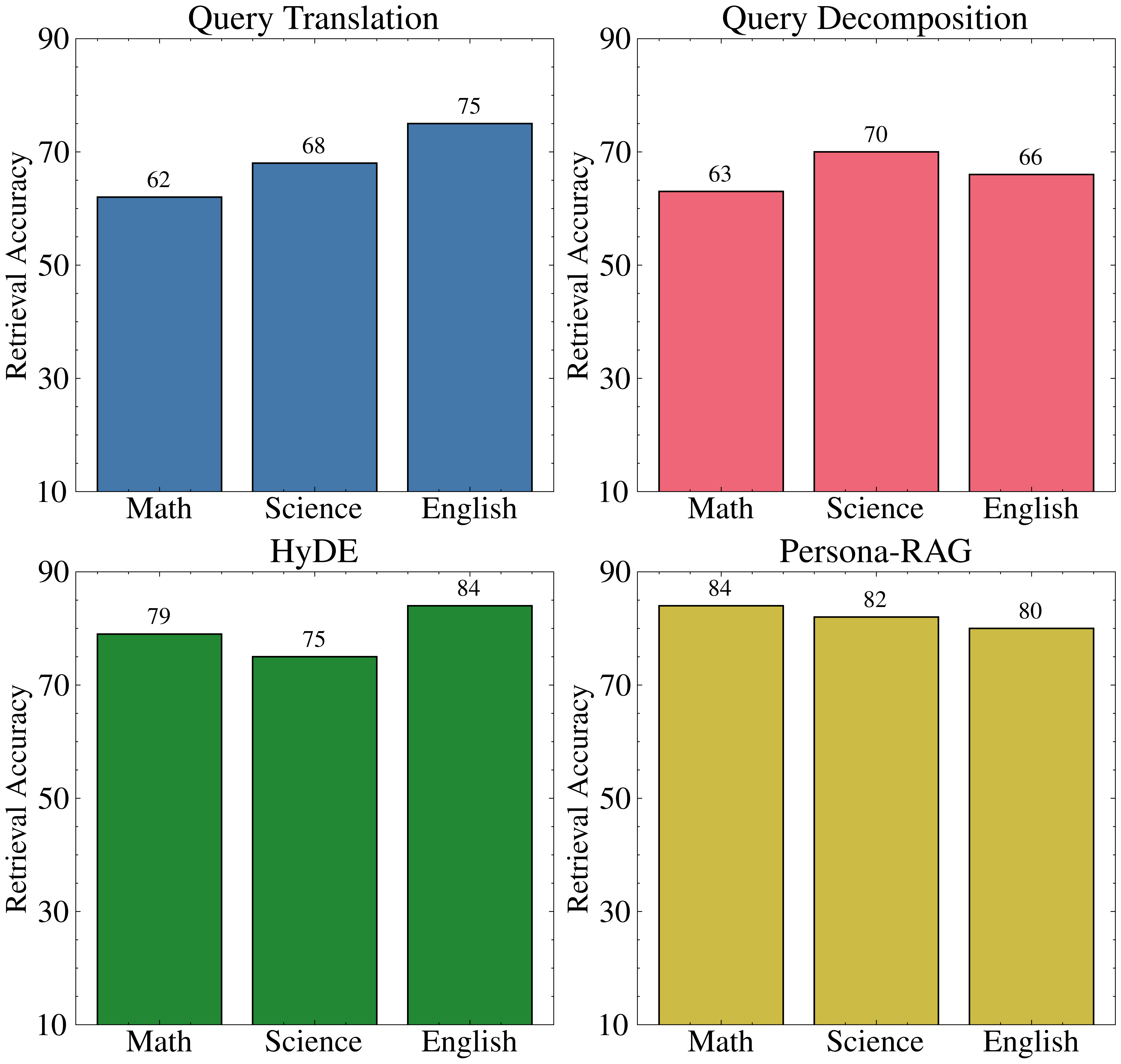}
     \caption{Subject-wise performance with different RAG methods. We observe that while Persona-RAG doesn't achieve the peak accuracies, the mean accuracy for each subject is high since it allows to student retrieve notes the in the same learning style as they wrote the notes in.}
     \label{fig:f1}
     \vspace{-1\baselineskip}
\end{figure}

\begin{figure}[t]
     \centering
     \includegraphics[width=0.45\textwidth]{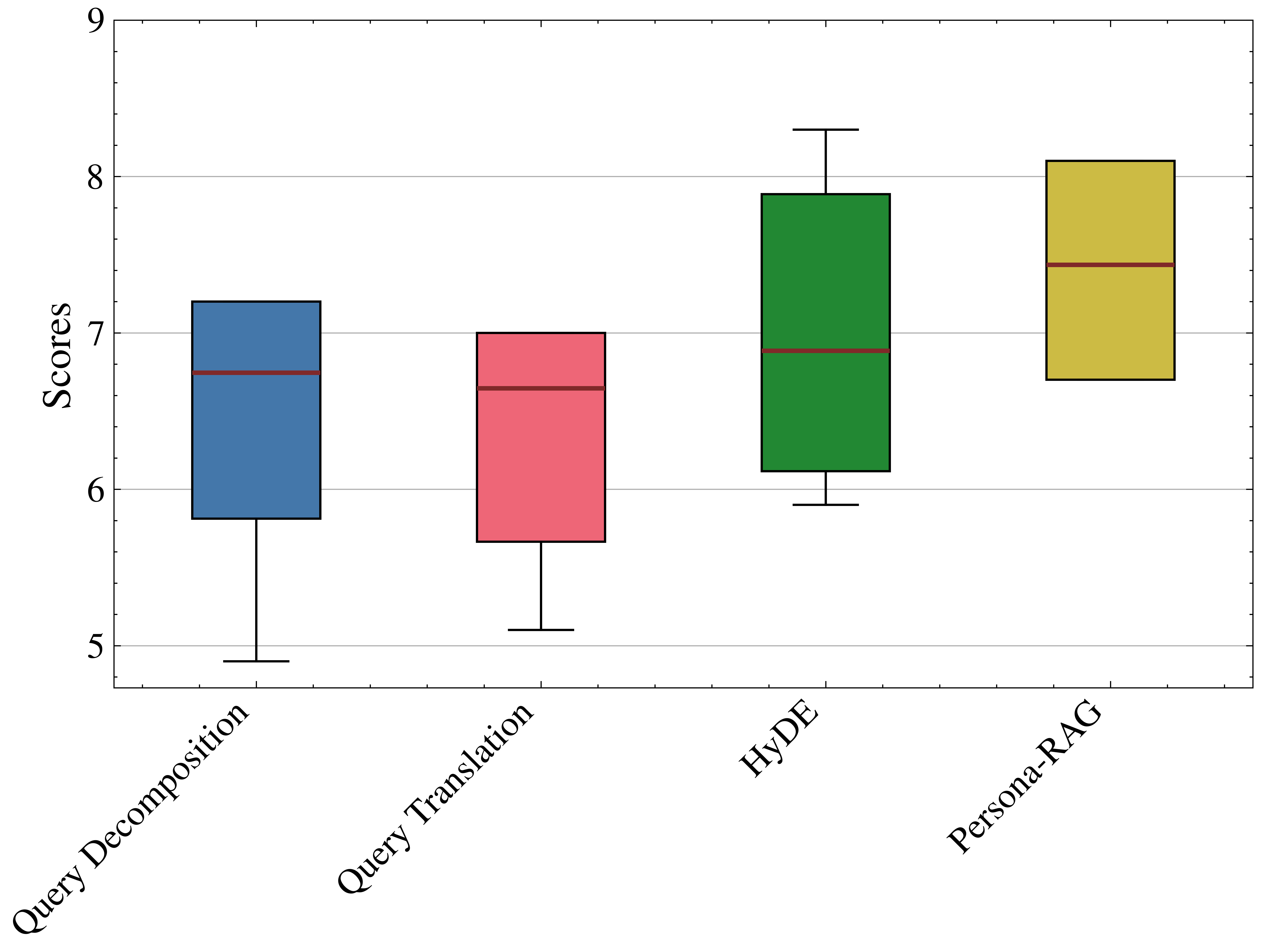}
     \caption{Box plots representing the distribution of student scores for different RAG techniques. While both Persona-RAG and HyDE have comparable performances, HyDE achieves the peak scores but Persona-RAG has a higher average score. It is worth noting that Persona-RAG is a better fit for pedagogical environment since it achieves comparative performance to HyDE at a fraction of time and compute.}
     \label{fig:f3}
     \vspace{-1\baselineskip}
\end{figure}

\subsection{Teacher Agent Pedagogical Strategy}
\label{sec:teach_strategy}
Effective teaching is inherently dynamic, requiring teachers to adapt their pedagogical strategies to resonate with students' varied learning styles and preferences. Simulating this adaptive process necessitates defining a controllable, yet sufficiently rich, set of teaching parameters that an automated agent can modify and optimize. While acknowledging that skilled teaching encompasses numerous nuanced factors often considered a craft, our goal is to operationalize key adjustable aspects of lecture delivery for computational simulation and optimization.
Simulating adaptive teaching requires defining controllable parameters for optimization. Based on consultation with experienced university professors to identify pivotal, modulable teaching dimensions impactful in diverse classrooms, we defined four major aspects of the teacher agent's pedagogical strategy:
\ding{182} \textbf{Explanation style}: \textit{How concepts are presented: technical, intuitive, visual, or auditory.}
\ding{183} \textbf{Content focus}: \textit{Analogies, real-world examples, or linking related concepts.}
\ding{184} \textbf{Delivery pace}: \textit{Slow or fast.}
\ding{185} \textbf{Engagement mode}: \textit{Lecture, example problems, or individual practice.}
These twelve variables are central to our framework because varying them allows the teacher agent to craft lecture deliveries specifically accommodating diverse student learning styles and personalities. For example, a visual learner benefits from slow, visual explanations with analogies and example problems, while a sequential/analytical learner may prefer fast, technical lectures with concept linking and individual practice.


\subsection{Persona-RAG}
We design a new Persona-RAG to be used for pedagogical teacher-student LLM agents framework. As given in Algorithm~\ref{alg:Persona-RAG}, for a question $Q$ and a student-trait vector $SC$, the student LLM first drafts a personalized reasoning plan $P=[p_1,\dots,p_m]$. Each step $p_i$ is then issued to \textit{Retrieve($p_i,KB$)}; the union of all returned chunks yields $D_{\text{retrieved}}$, the evidence pool for generation. Unlike standard query decomposition RAG which splits the \textit{question}, Persona-RAG splits the \textit{answer strategy}, letting evidence collection follow the learner’s own reasoning path. We discuss more details about Persona-RAG in the Appendix Section~\ref{sec:persona_rag}. 

\subsection{Teacher Agent Optimization via Genetic Algorithms}
\label{sec:gen_opt}
We move beyond observational studies of student-teacher interactions to actively optimize teaching patterns. The goal is to identify pedagogical strategies that maximize learning outcomes for a class comprising students with the diverse personalities and learning styles detailed in Section \ref{sec:intro}. This presents a complex optimization challenge. Each evaluation step within the optimization loop necessitates a full simulation cycle: \textit{the teacher agent delivers a lecture on a specific topic using a particular strategy configuration, student agents process this information and take personalized notes (updating their knowledge bases), students are assessed on the taught material, and their responses are evaluated to yield performance scores.}\par


Back propagating through the full pipeline: lecture generation, knowledge-base updates, Persona-RAG retrieval, student replies, and LLM-as-Judge scoring (Figure \ref{fig:pipeline}, Figure~\ref{fig:sample_results}) is prohibitively costly, so gradient methods would converge only after impractically long times. We instead adopt a Genetic Algorithm, which efficiently searches such expensive, non differentiable spaces. Each chromosome in the GA represents a teaching strategy configuration, defined by the pedagogical parameters from Section \ref{sec:teach_strategy}. The optimization process proceeds as follows:

\textbf{Step 1: Chromosome Encoding}: Each chromosome is a vector of numerical values corresponding to the chosen options and intensity for the four teaching variables.

\textbf{Step 2: Strategy Translation}: Values translate to natural language prompts for the teacher LLM using bucketing (e.g., 0.8 'Slow' -> "ensure pace is very slow").

\textbf{Step 3: Simulation \& Evaluation}: For each chromosome, the teacher lectures on a random topic. Student agents, based on style and personality, take personalized notes updating their KBs. Students are tested on 6 questions from the 3 subjects, the questions being a mix of the types detailed in Table \ref{tab:q_type}. Responses are rated 1-10 by Mistral-Large as an LLM-as-a-Judge.

\textbf{Step 4: Fitness Calculation}: Defined as the average assessment score of all student agents in the classroom for that lecture session.

\textbf{Step 5: Genetic Operations}: Based on their fitness, chromosomes are selected for reproduction using steady-state selection. Crossover and mutation have been applied to generate the next generation of teaching strategies, introducing variation and combining successful elements.

By iteratively evaluating strategies based on full simulation performance and selecting the fittest, the GA enables our teacher agent to discover pedagogical patterns incrementally improving diverse student learning outcomes. This framework is key to finding and analyzing adaptive teaching strategies in realistic, diverse LLM simulations.


\begin{table*}[]
    \centering
    \scriptsize
    \caption{Preferred teaching styles by students of varying learning styles denoted by average accuracies when taught with a specific teaching style. We observe how students from different learning styles have picked up on certain teaching patterns better, like \textit{Intuitive} learners prefer a lot of analogies with a fast-paced lecture. Similarly, \textit{Analytical} learners prefer technical lectures with a lot of interlinked related concepts to connect the patterns.}
    \label{tab:teach_type}
    \begin{tabular}{l|l|cccccc}
         \toprule
         \multicolumn{2}{l}{\textbf{Teaching Styles}} & \textbf{Read/Write} & \textbf{Visual} & \textbf{Auditory} & \textbf{Kinesthetic} & \textbf{Intuitive} & \textbf{Analytical} \\
         \midrule
          & Technical & 0.70 & 0.65 & 0.68 & 0.43 & 0.48 & \textbf{0.83} \\ 
         Explanation Style & Intuitive & 0.47 & 0.69 & 0.72 & 0.52 & \textbf{0.80} & 0.48 \\ 
         & Visual & 0.64 & \textbf{0.81} & 0.46 & 0.70 & 0.55 & 0.75 \\ 
         & Auditory & 0.50 & 0.49 & \textbf{0.78} & 0.50 & 0.72 & 0.52 \\ 
         \midrule
          & Analogies & 0.47 & 0.63 & 0.70 & 0.53 & \textbf{0.79} & 0.48 \\ 
         Content & Real-World Examples & 0.65 & \textbf{0.78} & 0.67 & 0.72 & 0.68 & 0.71 \\ 
         & Linking Related concepts & 0.60 & 0.62 & 0.61 & 0.58 & 0.69 & \textbf{0.76} \\ 
         \midrule
         {Delivery Pace} & Slow & 0.71 & 0.60 & 0.63 & 0.63 & 0.50 & \textbf{0.73} \\ 
         & Fast & 0.55 & 0.65 & 0.62 & 0.58 & \textbf{0.70} & 0.58 \\ 
         \midrule
          & Lecture & 0.65 & 0.55 & \textbf{0.78} & 0.45 & 0.70 & 0.68 \\ 
         Content Presentation & Example Problems & 0.58 & 0.70 & 0.47 & \textbf{0.80} & 0.65 & 0.72 \\ 
         & Individual Practise & \textbf{0.74} & 0.44 & 0.43 & 0.55 & 0.45 & 0.65 \\ 
         \bottomrule
    \end{tabular}
    \vspace{-1\baselineskip}
\end{table*}

\begin{figure*}[t]
     \centering
     \includegraphics[width=12cm, height = 7.8 cm]{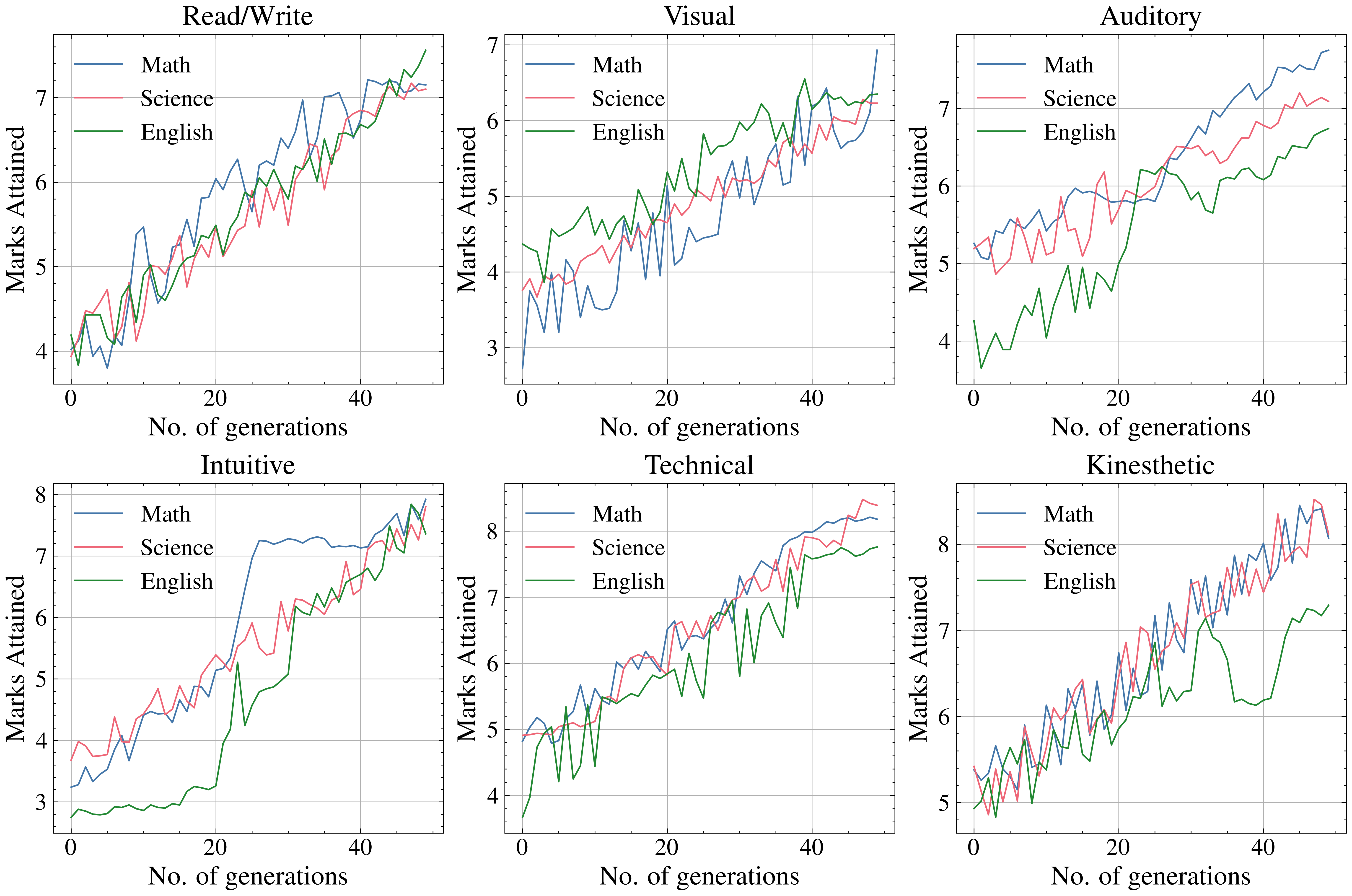}
     \caption{Performance of students with different learning styles over 50 generations. These experiments were done with the teacher optimizing over classes with students with homogeneous learning types. We observe a clear increasing trend in marks attained in each of the plots, indicating that the teacher can successfully identify the specific learning cues for each learning type. The peak average marks achieved by the class varies, and we can observe that \textit{Intuitive} and \textit{Technical} learners have higher average scores.}
     \label{fig:f2}
\end{figure*}

\begin{figure*}[t]
     \centering
     \includegraphics[width=15cm, height = 2.9 cm]{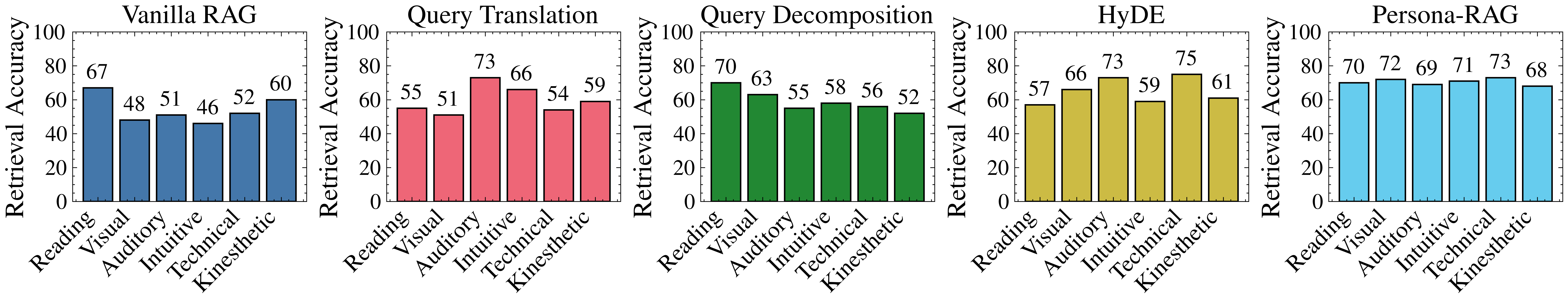}
     \caption{Comparing the retrieval accuracies of student with different learning styles with different RAG methods. A common trend that we observe is that most RAG methods are better for specific types of learning types, yielding high retrieval for them. HyDE works especially well with \textit{Technical} and \textit{Auditory} learners. Persona-RAG exhibits unbiased accuracies for all learning types, bringing out the best in them.}
     \label{fig:f4}
     \vspace{-1\baselineskip}
\end{figure*}

\section{Experiments and Results}
\label{sec:exp}
We evaluate our framework across two core modules: the retrieval quality and adaptability of Persona-RAG, and the effectiveness of our genetic algorithm (GA) in evolving pedagogical strategies for heterogeneous classrooms. Each simulation includes 20 student agents and one teacher agent using Mistral 3.1 Medium as backbone
\footnotetext{https://docs.mistral.ai/getting-started/models/models\_overview/\#premier-models}. Students maintain personal knowledge bases using FAISS HNSW (hierarchical navigable small world
graph)~\cite{douze2024faiss}~\cite{boytsov2016off}, and retrieval is used for assessment in Math, Science, and English.\par

\textbf{Persona-RAG achieves subject general retrieval accuracy with low computational overhead.}  
Figure~\ref{fig:f1} shows retrieval accuracy across subjects. Persona-RAG achieves the highest scores in Math (84\%) and Science (82\%), while maintaining strong performance in English (80\%). Although HyDE performs comparably in English, it requires generating a pseudo document for each query: impractical in real-time, multi-agent settings. Persona-RAG’s approach-based retrieval using 5–7 steps is significantly more efficient and easily parallelizable.\par

\textbf{Persona-RAG improves learning outcomes and retrieval stability.}  
Figure~\ref{fig:f3} displays student test score distributions under different retrieval methods. Persona-RAG results in the highest mean and lowest variance in scores, indicating that it not only improves average performance, but also reduces disparity among learners. This is especially important in pedagogical settings where consistency is as critical as peak performance.\par

\textbf{Persona-RAG is most effective on reasoning-based questions.} 
Table~\ref{tab:q_type} reports accuracy across five question types. While all methods perform similarly on Simple Recall, Persona-RAG outperforms others on Conceptual (88\%) and Analysis-Based (85\%) items, questions requiring abstraction, synthesis, and logical chaining. This supports the value of its plan-first retrieval design in aligning with high-order cognitive tasks.

\textbf{Genetic algorithm consistently improves classroom outcomes across generations.}  
Figure~\ref{fig:f2} presents average student scores over 50 GA generations, disaggregated by subject and learning style. A clear and monotonic increase in performance is observed across all subgroups, validating the GA’s ability to adapt teaching parameters via feedback. This confirms the viability of evolutionary optimization for non-differentiable, agent-based pedagogical modeling.\par

\textbf{Retrieval is robust across diverse cognitive styles.} 
Figure~\ref{fig:f4} shows retrieval accuracy per student learning style. Persona-RAG maintains consistent accuracy across all six styles, unlike baselines that show larger variances depending on student traits. This robustness highlights Persona-RAG's cognitive flexibility and suitability for heterogeneous learner populations.\par

\textbf{Learner trajectories reveal meaningful style-subject interactions.}  
From Figure~\ref{fig:f2}, intuitive learners show the highest gains, especially in Science. Visual and Auditory learners excel in Math, while Read/Write learners lead in English. Kinesthetic learners perform least well, suggesting mismatches between teaching modality and learning preference. These results provide interpretable diagnostic signals about group-level cognitive alignment.\par

\textbf{Style-specific preferences confirm pedagogical alignment.}  
Table~\ref{tab:teach_type} presents controlled evaluations of style-teaching compatibility. For instance, Intuitive learners perform best with analogies (0.79) and fast pacing (0.70), while Read/Write learners prefer technical explanation (0.70), slow delivery (0.71), and individual practice (0.74). The GA’s success in diverse classrooms suggests it implicitly integrates such preference signals over generations.\par

\textbf{Joint modeling of learning diversity and policy evolution enables generalizable and interpretable teaching strategies.} Together, these results demonstrate that combining cognitively grounded retrieval with adaptive policy evolution enables the emergence of robust, pedagogically sound, and interpretable teaching behaviors in simulated classrooms.

\section{Human Evaluation}
\label{sec:humaneval}
 To evaluate whether the strategies discovered by our framework exhibit real-world pedagogical value, we conducted a focused \textit{human evaluation}. We recruited a diverse group of 20 participants representing a spectrum of educational backgrounds: 9th-grade students, undergraduate students (freshmen and sophomores), and university professors. Each participant was shown lecture excerpts generated by our optimized teacher agent across three domains: Mathematics, Science, and English, covering introductory level topics. These lectures were produced using strategies evolved via our genetic algorithm, designed to optimize student engagement and conceptual clarity. Participants rated each lecture on a 1--10 scale along three key pedagogical dimensions: (1) \textbf{Approachability:} how accessible the content felt to a novice; (2) \textbf{Clarity of Core Ideas:} whether the central concepts were well explained; and (3) \textbf{Teaching Preference:} whether they would prefer this teaching style when first learning the topic. Table~\ref{tab:human_eval} presents the quantitative results. The average ratings are consistently high across domains: \textbf{Mathematics (8.45)}, \textbf{Science (8.65)}, and \textbf{English (8.05)}. These scores reflect a strong endorsement of the pedagogical quality of the lectures, suggesting that the teaching strategies optimized in simulation are not only effective computationally, but are also perceived as valuable by real learners and instructors.

To further contextualize these results, we analyzed the score distributions. Science exhibited a slightly higher standard deviation (2.08) compared to Math (1.41) and English (1.77), indicating more varied preferences. Qualitative feedback (see Appendix) suggests this variation stems from expectations about topic depth: some participants preferred a broad overview, while others desired more granular exposition. Importantly, even the 25th percentile ratings remain strong (Math: 7.5, Science: 7.5, English: 7.0), and median scores for all subjects fall at or above 8.0. \textit{These results serve as critical external validation: the learned teaching strategies are not only statistically sound but also pedagogically meaningful.} Participants consistently found the lectures clear, approachable, and well pitched, confirming that our simulation driven optimization produces teaching techniques with real world impact and promise for adaptive pedagogy.

\begin{table}[t]
    \centering
    \footnotesize
    \caption{Human rating statistics on lectures delivered by the teacher agent for all three subjects. We observe that humans prefer the intuitive style of lecture with lots of real-world examples, with a mean rating of 8 for all subjects and rating of 9 and above for the 75th percentile.}
    \label{tab:human_eval}
    \begin{tabular}{l|ccccc}
        \toprule
         \textbf{Subjects} & \textbf{Mean} & \textbf{Std. Dev.} & \textbf{P25} & \textbf{P50} & \textbf{P75}\\
         \midrule
         Mathematics & 8.45 & 1.41 & 6.5 & 8.0 & 9.5\\
         Science & 8.65 & 2.08 & 7.5 & 8.5 & 9\\
         English & 8.05 & 1.77 & 6.5 & 8.5 & 9\\
         \bottomrule
    \end{tabular}
    \vspace{-2\baselineskip}
\end{table}

\section{Conclusion}
We address the critical challenge of discovering adaptive pedagogical strategies tailored to diverse student populations, advancing beyond traditional, static simulation paradigms. We present a dynamic classroom simulation where LLM-based student agents learn from a GA-evolved teacher. The genetic search uncovers pedagogical parameters that reliably boost student performance and well suited for complex, non-differentiable interaction landscapes. Complementing this,~\textit{Persona-RAG} personalizes retrieval to each student and surpasses standard RAG on complex queries while staying robust across profiles. Human judges deem the discovered strategies both plausible and educationally sound. These contributions form a testbed for AI-driven adaptive teaching and lay a foundation for scalable, feedback-driven teacher-training tools.


\section{Limitations}
\label{sec:limitations}
While our framework demonstrates promising results in optimizing adaptive teaching strategies and validates their effectiveness through human evaluation, several limitations remain that open avenues for future exploration. First, due to computational constraints, the simulated curriculum was limited in scope. Although carefully designed to be representative, it does not span the full breadth of subjects and cognitive skills typical of a pre-undergraduate syllabus. Extending the curriculum to cover a wider range of topics would allow for deeper analysis of the generalizability and robustness of learned teaching strategies. Second, our student modeling incorporates key dimensions such as learning styles and personality traits, but remains coarse grained. Capturing finer grained characteristics: such as cognitive development stages, affective states, or temporal learning trajectories could yield more realistic simulations and support richer pedagogical adaptations. Lastly, while our study establishes the viability of multi-agent simulation for personalized education, it remains at a mid-scale experimental level. Future work could expand both the number and diversity of agents, and explore real world integration scenarios, such as intelligent tutoring systems or teacher-assistive platforms. In sum, our framework serves as a strong foundation for future research in AI-driven education. By expanding curriculum coverage, refining student representations, and scaling up simulation complexity, the community can move closer to building realistic, adaptive, and equitable educational technologies.

\bibliography{main_bib.bib}

\newpage

\begin{table*}[]
    \centering
    \footnotesize
    \caption{Increasing the number of student agents and evaluating its impact on the proposed framework}
    \label{tab:class_size_ablation}
    \begin{tabular}{c|ccccccc}
    \toprule
        \textbf{No. of students} & \textbf{Mean} & \textbf{Std. Dev.} & \textbf{P25} & \textbf{P50} & \textbf{P75} & \textbf{90\% Plateau} & \textbf{Time in minutes} \\
        \midrule
         10 & 7.40 & 2.03 & 5.81 & 6.92 & 8.08 & 32 & 44 \\
         25 & 8.33 & 1.94 & 7.09 & 8.39 & 9.61 & 51 & 97 \\
         50 & 8.28 & 1.77 & 7.04 & 8.25 & 9.46 & 67 & 203 \\
         75 & 7.91 & 1.70 & 7.01 & 8.18 & 9.29 & 82 & 288 \\
         100 & 8.14 & 1.62 & 7.05 & 8.02 & 9.00 & 86 & 391 \\
         \bottomrule
    \end{tabular}
\end{table*}

\section{Appendix A: Additional Study}

\begin{figure*}[t]
     \centering
     \includegraphics[width=12cm, height = 7.8 cm]{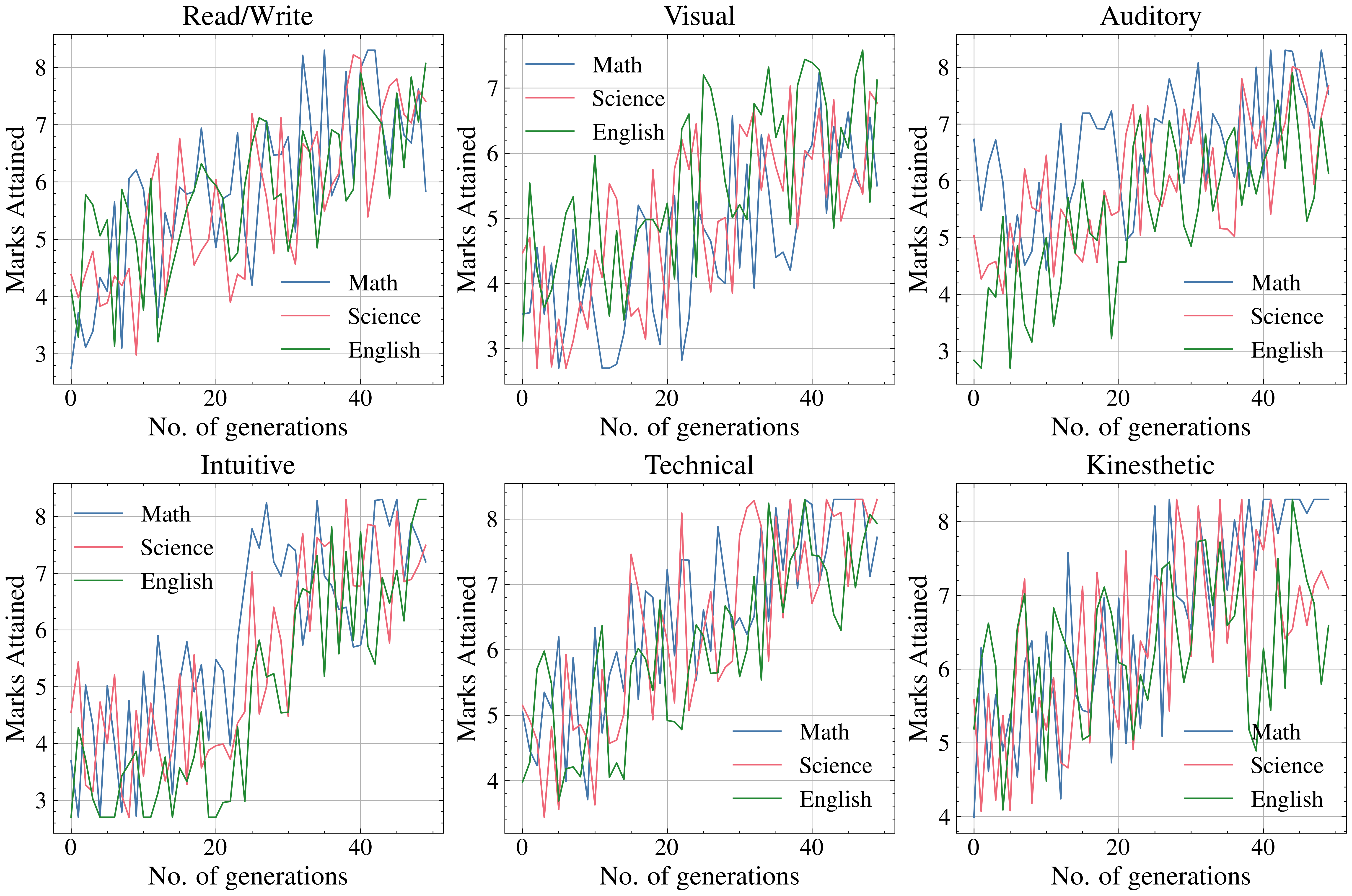}
     \caption{Random selection in GA: Performance of students with different learning styles over 50 generations}
     \label{fig:selection_random}
     \vspace{-1\baselineskip}
\end{figure*}

\begin{figure*}[t]
     \centering
     \includegraphics[width=12cm, height = 7.8 cm]{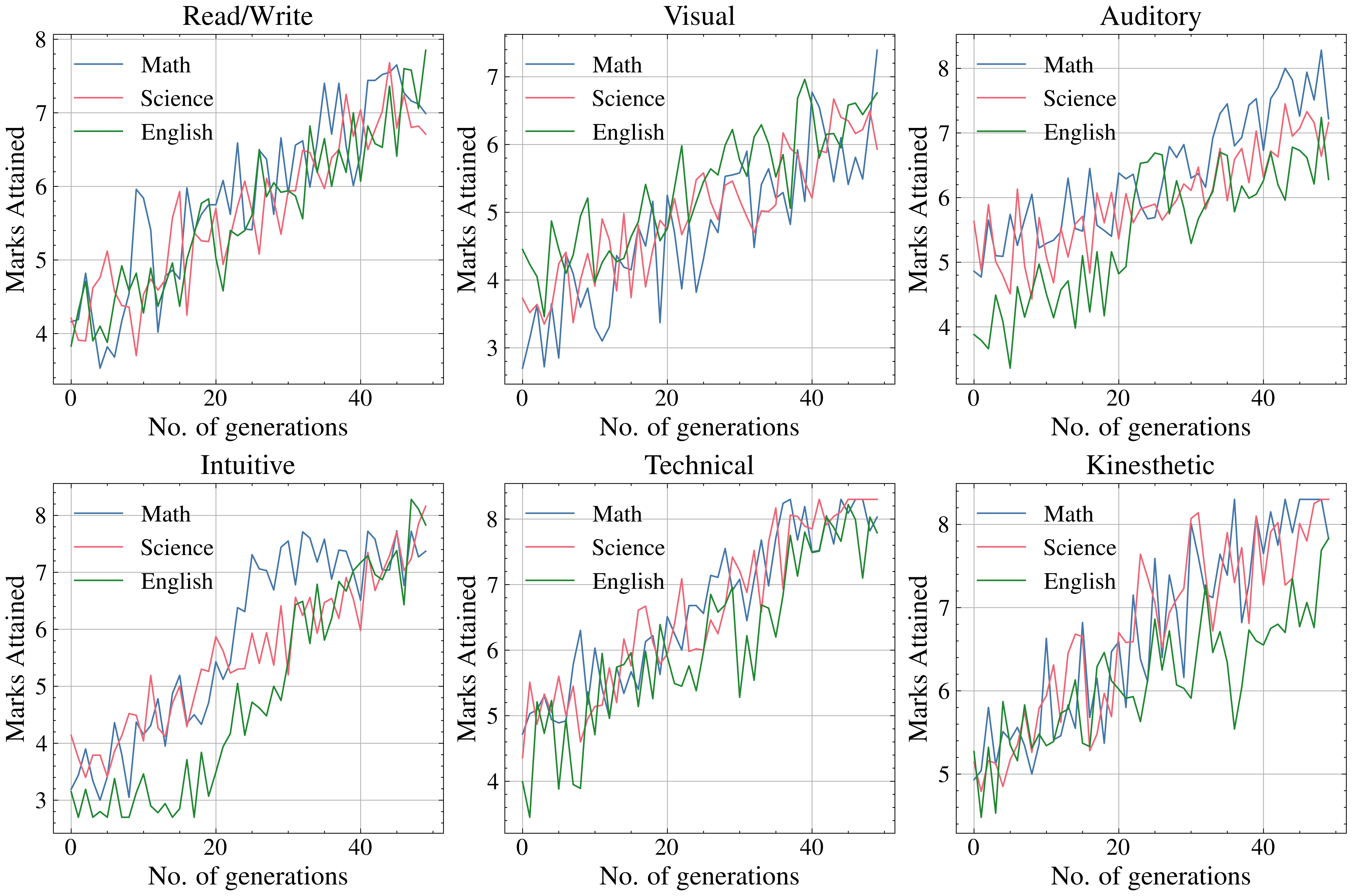}
     \caption{Tournament selection in GA: Performance of students with different learning styles over 50 generations}
     \label{fig:selection_tournament}
     \vspace{-1\baselineskip}
\end{figure*}

\begin{figure*}[t]
     \centering
     \includegraphics[width=12cm, height = 7.8 cm]{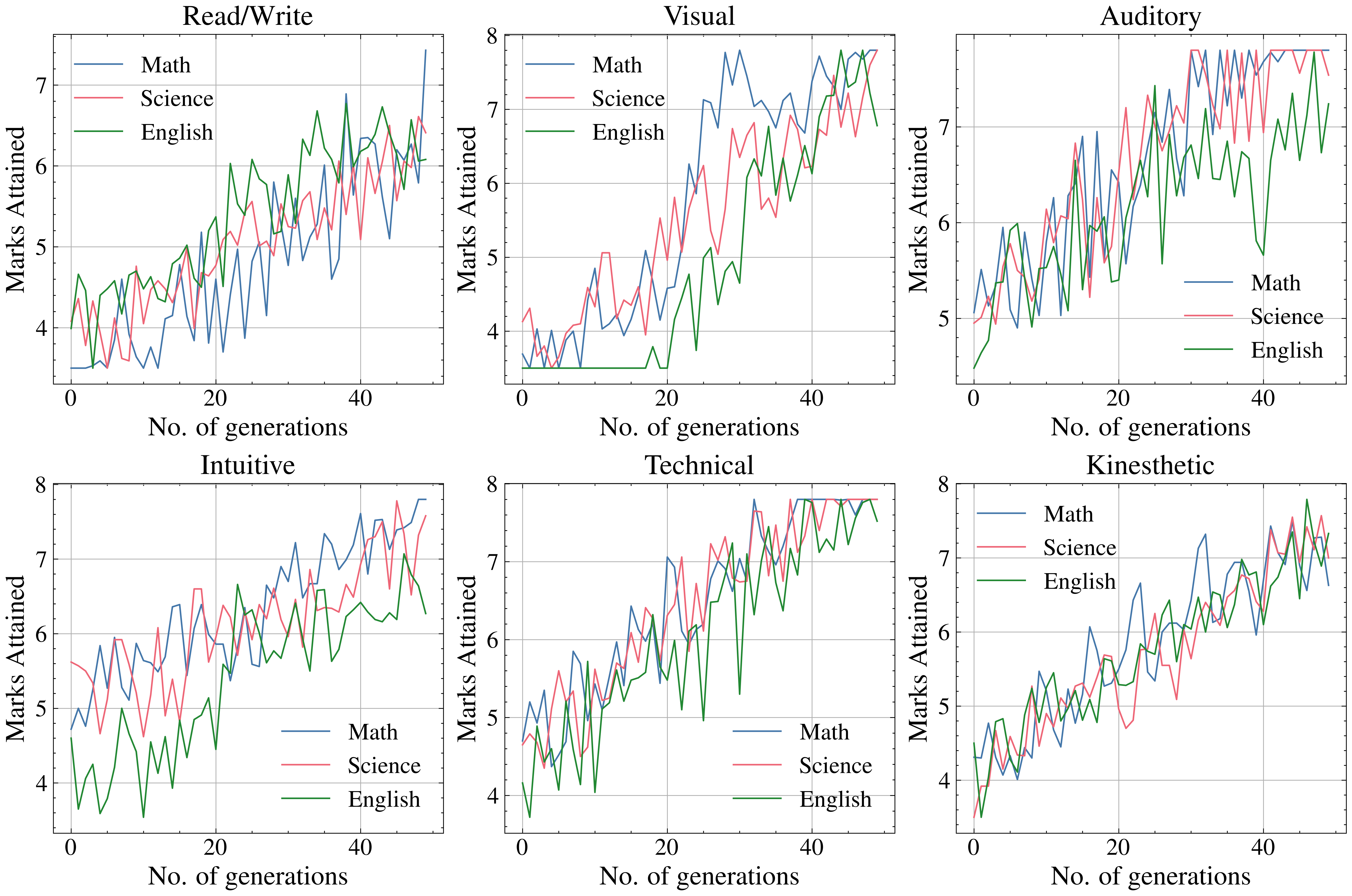}
     \caption{Roulette wheel selection in GA: Performance of students with different learning styles over 50 generations}
     \label{fig:selection_roulette}
     \vspace{-1\baselineskip}
\end{figure*}
\subsection{Scalability of Teacher Optimization with Increasing Class Size.}  
We investigate how the teacher agent’s optimization dynamics and the quality of the resulting pedagogical strategies scale with the number of students in the classroom. This ablation study evaluates our framework’s robustness under increasing class sizes, from 10 to 100 student agents, and summarizes the results in Table~\ref{tab:class_size_ablation}. Across all evaluated sizes, the genetic algorithm (GA) consistently converges to high-performing teaching strategies. The final mean student score ranges from 7.40 (10 students) to 8.14 (100 students), with peak performance observed between 25 and 50 students. This indicates that the GA effectively aggregates feedback and discovers generalizable strategies even in increasingly diverse and complex student populations.

Crucially, score distribution metrics remain stable as class size increases. The 25th percentile consistently exceeds 7.0 for classes with 25 or more students, suggesting that performance at the lower end of the distribution does not degrade with scale. Moreover, the standard deviation of scores exhibits a mild decreasing trend: from 2.03 (10 students) to 1.62 (100 students), implying that optimization over larger populations results in pedagogical policies that are more balanced and less sensitive to individual variability. These findings highlight the GA’s capacity to optimize for collective performance without sacrificing equity. As expected, scalability introduces computational costs. The number of generations required to reach 90\% of the final performance increases from 32 (10 students) to 86 (100 students), as shown in Table~\ref{tab:class_size_ablation}. This growth reflects the broader feedback landscape encountered when optimizing over more agents. Similarly, the per-generation runtime scales approximately linearly, from 44 minutes (10 students) to 6.5 hours (100 students), due to the increasing computational load from student-specific note-taking, Persona-RAG retrieval, and assessment. Despite this, the framework remains tractable for class sizes that reflect realistic educational settings. Overall, these results demonstrate that our framework maintains pedagogical effectiveness and fairness as classroom size increases, while scaling computationally in a predictable and interpretable manner.

\subsection{Different Types of Parent Selection in GA}
The parent selection mechanism is a fundamental component of Genetic Algorithm, governing how individuals from the current generation are chosen to create the next. The efficacy of selection directly impacts the algorithm's ability to converge on optimal solutions by balancing the exploitation of high-performing strategies with the exploration of the search space. To understand the critical role of selection in optimizing our teacher agent's pedagogical strategy, we conducted an ablation study comparing the optimization performance using three common parent selection types against the Steady-State selection employed in our main experiments: Random Selection, Tournament Selection, and Roulette Wheel Selection. The learning curves, showing the average student scores over 50 generations for each learning style and subject, are presented in Figure \ref{fig:selection_random}, Figure \ref{fig:selection_tournament}, and Figure \ref{fig:selection_roulette} for Random, Tournament, and Roulette Wheel selection, respectively.

Our analysis confirms that an effective parent selection mechanism is crucial for the Genetic Algorithm to successfully optimize teacher behavior in our simulation. As depicted in Figure \ref{fig:selection_random}, employing Random Selection for parent generation results in negligible optimization progress. The average student scores across all learning styles and subjects exhibit high volatility and no discernible upward trend over 50 generations, remaining close to the initial performance levels. This outcome is intuitive: Random Selection provides no evolutionary pressure, failing to favor fitter individuals and thus reducing the GA to a random search process ineffective for navigating the complex, high-dimensional strategy space of our teacher agent.\par

In contrast, selection methods that introduce evolutionary pressure demonstrate clear optimization capabilities, driving significant improvements in student learning outcomes. Both Tournament Selection (Figure \ref{fig:selection_tournament}) and Roulette Wheel Selection (Figure \ref{fig:selection_roulette}) show consistent and substantial increases in average student scores across all learning styles and subjects over the 50 generations. By favoring chromosomes with higher fitness (better average class scores), these methods effectively propagate successful pedagogical strategies through the population, guiding the search towards more optimal teaching patterns. While both methods facilitate optimization, they differ in their balance of exploitation and exploration; Tournament Selection is typically less sensitive to outlier fitness values, potentially leading to steadier convergence, whereas Roulette Wheel selection's probabilistic nature proportional to fitness can be more susceptible to super-fit individuals dominating the population early. The observed clear learning curves with these methods highlight that introducing a mechanism to select and recombine higher-performing strategies is essential for the GA to discover effective adaptive teaching behaviors in our environment.\par

The results from this ablation study underscore the critical role of the parent selection mechanism in enabling the successful optimization of our teacher agent. They validate that a selection strategy providing adequate evolutionary pressure is necessary for the Genetic Algorithm to effectively navigate the pedagogical strategy space and discover teaching patterns that lead to improved student learning outcomes. This empirical evidence justifies our choice of a robust selection method (Steady-State selection, as used in the main results in Section 4) as a foundational component for achieving successful teacher adaptation in our diverse multi-agent simulation.

\begin{figure*}[t]
     \centering
     \includegraphics[width=10cm, height = 5.8 cm]{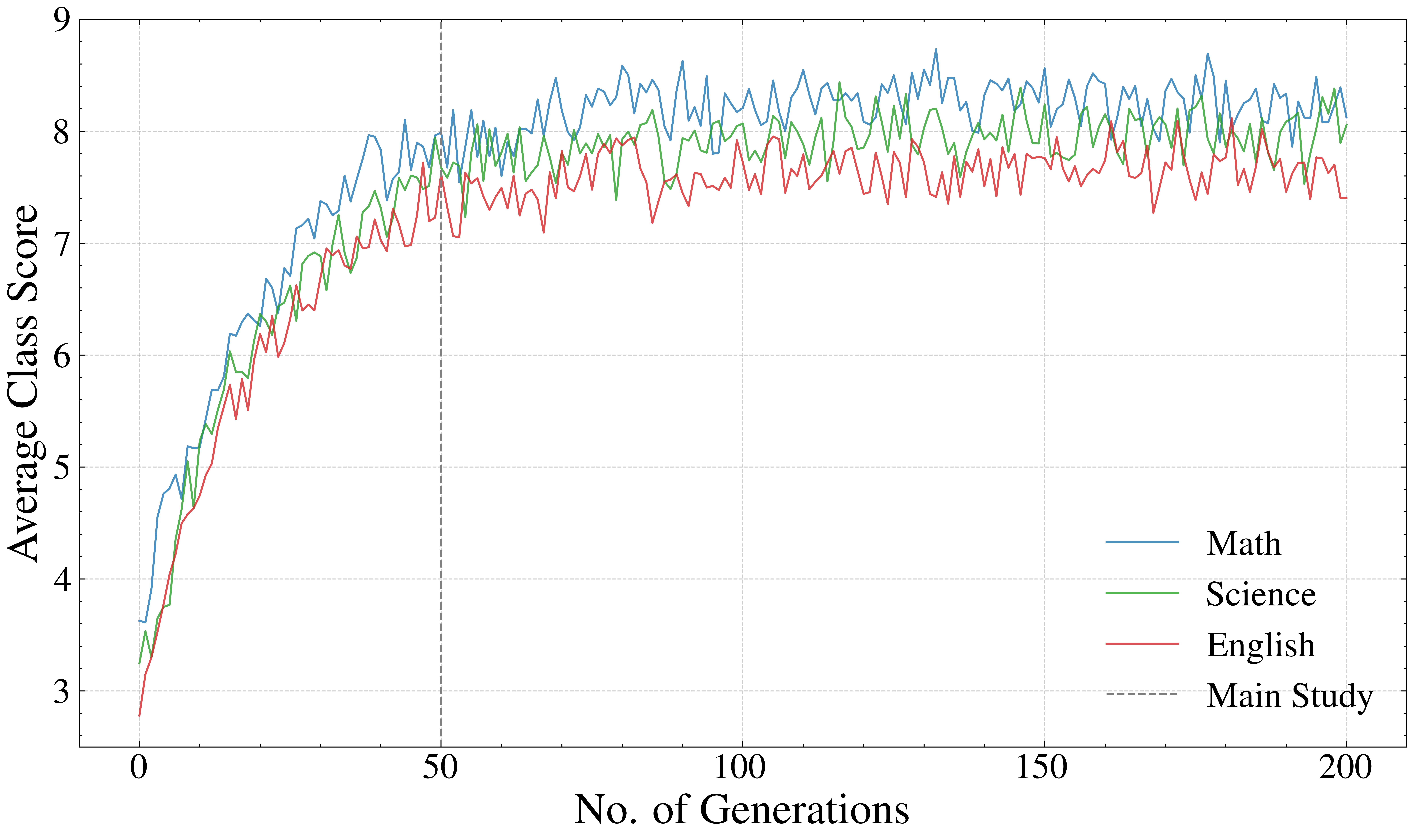}
     \caption{Convergence of the Genetic Algorithm teacher optimization. Curves show the average class score for Math, Science, and English across 200 generations; the dashed grey line marks the 50-generation horizon used in main experiments. Scores rise steeply from the random baseline to their plateau by $\sim$70 generations, after which improvements are marginal. The plateau confirms that 50 generations already capture the bulk of attainable gains, while further iterations offer diminishing returns.}
     \label{fig:extended_generations}
     \vspace{-1\baselineskip}
\end{figure*}

\begin{table}[]
    \centering
    \small
    \resizebox{\linewidth}{!}{%
    \begin{tabular}{l|l}
         \toprule
         \textbf{Subject} & \textbf{Topic} \\
         \midrule
         & Simplifying Expressions \\
         & Linear Equations and Inequalities \\
         & Working with Formulas \\
         & Functions Concept \\
         & Cartesian Coordinate System \\
         & Distance Formula \\
         & Midpoint Formula \\
         & Slope of a Line \\
         & Equations of Lines \\
         Math & Parallel and Perpendicular Lines \\
         & Triangles \\
         & Right Triangles and Pythagorean Theorem \\
         & Angles and Angle Relationships \\
         & Circles Basic \\
         & Similar Figures and Similar Triangles \\
         & Congruent Figures Basic \\
         & Number Sense and Operations \\
         & Data Handling and Basic Statistics \\
         & Logical Reasoning and Problem Solving \\
         \midrule
         & Motion Basics \\
        & Acceleration \\
        & Units of Measurement Physics \\
        & Motion Graphs Introductory \\
        & Ratio and Proportion \\
        & Atomic Structure Basics \\
        & Elements and Periodic Table Introductory \\
        & Electrons and Electron Shells Simplified \\
        & Electric Charge Basic \\
        Science & Ions Introductory \\
        & Cells as the Basic Unit of Life \\
        & Prokaryotic and Eukaryotic Cells \\
        & Basic Cell Processes Overview \\
        & Respiration Energy Use \\
        & Growth and Reproduction \\
        & Response to Environment \\
        & Nutrient Uptake \\
        & Waste Removal \\
        & Molecules and Macromolecules Basic Idea \\
        & Water and Its Importance \\
        \midrule
         & Shakespearean Drama \\
          & Poetry \\
          & Short Story \\
          & Identifying Explicit Information \\
          & Following Plot and Sequence of Events \\
          & Character Identification \\
          & Drawing Inferences \\
          & Recognizing Basic Symbolism and Figurative Language  \\
          & Plot Elements \\
          & Character Types \\
          & Setting Importance \\
          English & Drama/Play  \\
          & Simile and Metaphor \\
          & Imagery \\
          & Personification \\
          & Understanding Varied Sentence Structures \\
          & Ability to Follow Complex Texts \\
          & Identifying Author's Purpose \\
          & Supporting Opinions with Textual Evidence \\
          & Making Connections \\
          & Drama Basics \\
          & Historical Context \\
          & Figurative Language \\
          & Sound Devices Awareness \\
          & Narrative Structure \\
         \bottomrule
    \end{tabular}}
    \caption{Prerequisite list of topics for Students}
    \label{tab:prerequisites}
\end{table}

\begin{table}[]
    \centering
    \footnotesize
    \resizebox{\linewidth}{!}{%
    \begin{tabular}{l|l|l}
         \toprule
         \textbf{Subject} & \textbf{Topic} & \textbf{Subtopic}\\
         \midrule
          & & Trigonometric Ratios \\
          & Trigonometry & Trigonometric Identities \\
          & & Applications in Heights and Distances \\
          & & Basic concepts\\
          Mathematics & Probability & Simple and Compound events\\
          & & Complimentary events\\
          & & circles\\
          & Analytical Geometry& Ellipses\\
          & & hyperbolas\\
          \midrule
          & & newtons first law\\
          & Newton's Laws & newton, second law\\
          & & newtons third law\\
          & & ionic bonding    \\
          Science & Chemical Bonding & covalent bonding            \\
          & & valence, and Lewis structures \\            
          & & structure of prokaryotic and eukaryotic cells            \\
          & Cell Structure & function of cell organism            \\
          & & difference between plant and animal cell            \\
          \midrule
          & & themes in Shakespeare in place\\
          & Shakespearean Drama & character analysis\\
          English & & dramatic devices\\
          & & figurative language\\
          & Poetry & themes, and tone\\
          & & poetic devices and structure\\
          \bottomrule
    \end{tabular}
    }
    \caption{List of topics to teach the students}
    \label{tab:curriculum}
\end{table}

\subsection{Increasing the Number of Generations}
The number of generations for which a Genetic Algorithm is run is a critical parameter, directly impacting the trade-off between computational cost and the extent to which the optimization explores the solution space. Running for too few generations may result in premature convergence or failure to find near-optimal solutions, while running for excessive generations is computationally wasteful if performance has already plateaued. This ablation study investigates the convergence dynamics of our teacher agent optimization by extending the number of generations significantly beyond the 50 generations used in our main experiments. Figure \ref{fig:extended_generations} presents the average class score for each subject (Math, Science, English) over 200 generations of the Genetic Algorithm.\par

The Genetic Algorithm exhibits rapid initial learning followed by a clear plateau, demonstrating reliable convergence of the teacher optimization process. As shown in Figure \ref{fig:extended_generations}, the average class scores for Math, Science, and English increase sharply from their initial low values (around 3.0-3.5) up to approximately generations 50-70. This initial steep rise reflects the GA quickly identifying and propagating teaching strategies that are significantly more effective than the random starting points, validating its ability to find beneficial pedagogical patterns early in the optimization process. Beyond this initial phase, the learning curves for all three subjects show a distinct flattening, indicating that the average scores have largely plateaued. While minor fluctuations persist due to the stochastic nature of the simulation and GA operations, there is no sustained upward trend in performance from around generation 70 onwards up to 200 generations. This indicates that the GA has converged to near-optimal teacher strategies within the defined pedagogical parameter space.
The observed plateauing validates that running the optimization for 50 generations, as in our main study, captures the majority of the achievable performance gains. The dashed grey line at Generation 50 in Figure \ref{fig:extended_generations} clearly shows that by the end of our main experimental run, the average class scores for all subjects had already reached a significant portion of their final plateaued value. While extending the optimization further might yield marginal refinements, this ablation demonstrates that the computational resources invested in 50 generations provide a strong return in terms of finding highly effective teaching strategies. Running for significantly longer periods, while confirming stability, offers diminishing returns in terms of average performance improvement.\par

This ablation study provides crucial evidence for the convergence properties and efficiency of our GA-based teacher optimization. It confirms that the algorithm effectively explores the strategy space and finds near-optimal solutions within a practical number of generations, justifying the experimental setup used throughout our study and reinforcing the reliability of the learned adaptive teaching patterns.

\subsection{Prerequisites for Students}
Creating a realistic simulated classroom environment necessitates endowing student agents with foundational knowledge appropriate for the curriculum they will be taught. Just as human students enter classrooms with varying levels of prior understanding, our LLM agents require a defined prerequisite knowledge base to engage meaningfully with the teacher agent's lectures and complex assignments. This is particularly critical for tasks requiring analytical reasoning and problem-solving, where a solid grasp of fundamental concepts is essential for building up to more complex applications.
To ensure our student agents are equipped with relevant foundational understanding and to provide a realistic starting point for simulating diverse prior aptitudes, we carefully curated a prerequisite knowledge base. The selection of topics was informed by consultation with a panel of experienced university professors, identifying concepts deemed essential for pre-undergraduate level comprehension in Mathematics, Science, and English. This expert guidance helped us identify the core building blocks necessary for students to approach new material from a ground-up perspective, particularly enabling their ability to tackle analytical and derivation-based questions.\par

The prerequisite knowledge base was constructed by drawing topics from widely recognized international pre-university examination syllabi, including SAT, JEE, and Gaokao. This grounding in established curricula ensures that the foundational knowledge within our simulation aligns with common educational benchmarks. As detailed in Table \ref{tab:prerequisites}, the knowledge base comprises a comprehensive list of topics: 20 topics in Mathematics (primarily covering foundational Algebra and Geometry), 21 topics in Science (spanning core concepts in Physics, Chemistry, and Biology), and 37 topics related to essential English Literature and Grammar principles. Furthermore, as described in the main paper, to realistically model students with variable subject aptitudes, the knowledge for each of these prerequisite topics is structured across three distinct levels of detail (Level 1: overview, Level 2: foundations, Level 3: in-depth). This layered representation allows us to initialize student agents with differing depths of understanding for the same topic depending on their simulated aptitude for that subject, moving beyond a uniform baseline and contributing significantly to the diversity of our simulated student population.
In summary, the prerequisite knowledge base serves as a vital component of our simulation framework, providing a standardized yet customizable foundation for student learning. Its construction, guided by expert input and standard curricula, and its multi-level structure are designed to support realistic student modeling and enable effective evaluation of adaptive teaching strategies for diverse learners.

\subsection{Topics to Teach}
The core function of our teacher agent is to deliver lectures on specific topics within a defined curriculum, enabling the simulation of the learning process and the subsequent evaluation and optimization of teaching strategies. The selection of this curriculum is therefore integral to our framework, providing the environment in which student-teacher interactions occur and student learning is measured.
Building directly upon the prerequisite knowledge described in Appendix A, the curriculum taught by the teacher agent comprises subsequent concepts in Mathematics, Science, and English. These topics were carefully chosen to require the application and extension of the foundational concepts present in the prerequisite knowledge base. The intention behind this design is to simulate a realistic learning progression where students must leverage their prior understanding to grasp new, more advanced material, thereby allowing us to assess their ability to build knowledge and apply foundational principles. This structure is particularly important for evaluating student performance on question types that demand analytical thinking and synthesis, which rely heavily on a solid grasp of prerequisite building blocks.\par

Furthermore, this specific curriculum was selected for its suitability in testing the capabilities of our framework, including the adaptive teacher and Persona-RAG. The topics include concepts that can be approached and explained using diverse pedagogical methods (e.g., deriving formulas in Math, explaining biological processes in Science, analyzing literary devices in English), providing a rich space for the teacher agent's parameterized strategies (Section 3.2) to explore and adapt. Crucially, these topics naturally lend themselves to generating a variety of assessment questions, encompassing simple recall, conceptual understanding, application, analysis, and creative tasks (as discussed in Section 4). This diversity in question types is essential for a comprehensive evaluation of student learning and, specifically, for demonstrating the effectiveness of Persona-RAG on tasks beyond simple factual retrieval.\par

As detailed in Table \ref{tab:curriculum}, the curriculum includes key topics such as Trigonometry, Probability, and Analytical Geometry in Mathematics; Newton's Laws, Chemical Bonding, and Cell Structure in Science; and Shakespearean Drama and Poetry in English. This selection ensures a breadth of concepts that challenge students and provide varied contexts for teaching and learning within our simulation.
In summary, the curriculum taught by the teacher agent is strategically designed to build upon the prerequisite knowledge, provide a rich environment for exploring adaptive teaching strategies, and support the generation of diverse assessment questions necessary for rigorous evaluation of both student learning and the effectiveness of our Persona-RAG method.

\section{Appendix B: Personalized Knowledge Retrieval (Persona-RAG)}
\label{sec:persona_rag} 
Effective knowledge retrieval from their internal knowledge bases is a critical capability for our student agents, significantly contributing to the fidelity of the simulation and the accuracy of their test performance. During assessments, students must access and synthesize relevant information learned from lectures and their prerequisite knowledge to construct answers. However, integrating realistic and effective information retrieval into diverse LLM agents for complex pedagogical tasks presents unique challenges not fully met by standard Retrieval-Augmented Generation (RAG) approaches.

We identified two key limitations of existing RAG methods within the context of our pedagogical simulation. First, empirically, we observed that traditional RAG approaches such as Query Translation, Query Decomposition~\cite{chan2024rqraglearningrefinequeries}, HyDE~\cite{gao2022precisezeroshotdenseretrieval}, and similar methods often exhibit decreased performance when faced with non-recall based conceptual or analysis questions. These tasks, requiring multi-step reasoning, synthesis, or derivation rather than simple factual lookup, demand a more structured approach to information retrieval that current query-centric methods struggle to fully capture. Second, existing RAG techniques are predominantly query-dependent, focusing solely on transforming or enriching the input query itself. They lack a mechanism to incorporate individual agent characteristics into the retrieval process. Real students exhibit diverse aptitudes and strategies for recalling and organizing information based on their cognitive styles and personalities, factors which profoundly shape their ability to perform during assessments. A realistic simulation requires a retrieval mechanism that can reflect these individual differences, moving beyond a monolithic, query-only retrieval model.

To address these limitations and enable a more realistic and personalized approach to knowledge retrieval for our student agents, we introduce \textbf{Persona-RAG}. Inspired by human cognitive processes when tackling complex questions, Persona-RAG operates on the principle that a student, confronted with a non-recall question, first forms a mental structure or strategy for approaching the problem and organizing the required information before retrieving specific details. This personalized structuring process provides a crucial context for subsequent information lookup and synthesis.

In Persona-RAG, given an input question $Q \in \mathcal{S}$ (where $\mathcal{S}$ is the set of strings) and a student agent with specific characteristics $SC$ (a structured representation encoding cognitive style, memory preferences, and learning traits), the process unfolds in two main stages, as outlined in Algorithm~\ref{alg:Persona-RAG}. First, the student agent generates a personalized approach structure $P = [p_1, \dots, p_m]$ for answering the question $Q$. This structure is a sequence of $m$ abstract cognitive elements or key points required to construct a complete answer. Crucially, the generation of this structure $P$ is influenced by the student's characteristics $SC$. This personalization allows the structure and focus of the approach to differ across students even for the same question $Q$, capturing individual strategies for problem-solving and information organization. We formalize this stage as: $P = \text{PERSONAGEN\_STRUCTURE}(Q, SC)$, where $\text{PERSONAGEN\_STRUCTURE}$ is a function (implemented using the student's LLM agent) that takes the question $Q$ and student characteristics $SC$ as input and outputs a list of structure elements $p_i \in \mathcal{S}$.

Second, instead of relying on a single query derived directly from $Q$, Persona-RAG leverages the generated approach structure $P$. For each element $p_i$ in the structure $P$, retrieval is performed directly using $p_i$ as the input to the retrieval function $\text{RETRIEVE}$. The $\text{RETRIEVE}$ function queries the student's knowledge base $KB = \{d_1, \dots, d_n\}$ (a corpus of $n$ indexed document chunks, where $d_j \in \mathcal{S}$) and returns a ranked subset of relevant document chunks $D_i \subseteq KB$ that semantically match the structure element $p_i$. The retrieved document sets $D_i$ from each structure element are then aggregated to form the final set of retrieved documents $D_{\text{retrieved}}$. We formalize this stage as: For $i = 1$ to $m$: $D_i = \text{RETRIEVE}(p_i, KB)$, and $D_{\text{retrieved}} = D_{\text{retrieved}} \cup D_i$, where $D_{\text{retrieved}}$ is initialized as an empty set.

It is important to distinguish Persona-RAG from Query Decomposition~\cite{chan2024rqraglearningrefinequeries}. While Query Decomposition breaks down the input \textit{question} into simpler sub-questions to facilitate retrieval, Persona-RAG breaks down the \textit{approach to solving the question} or the \textit{structure of the intended answer} into personalized elements based on the student's traits. This difference in focus allows Persona-RAG to provide a more robust framework for retrieving information needed for multi-step reasoning and synthesis, and crucially, enables the personalization of the retrieval process based on student-specific strategies generated from their characteristics.

\clearpage
\clearpage
\onecolumn

\begin{tcolorbox}[
enhanced jigsaw,
height=\textheight,
size=small,
colframe=blue!50!white,
colback=white!10!white,
before title=\raggedright,
title={\textbf{Lecture Plan: Applications in Heights and Distances}},
fonttitle=\bfseries,
pad at break=1mm
]
{
\fontsize{10}{12}\selectfont
\section*{} 

\subsection*{\textcolor{violet}{\textbf{1. Introduction (5 minutes)}}} \textbf{Objective:} Hook students with a real-world scenario and set the stage for the lesson.

\textbf{Approach:} Start with a friendly, conversational tone: \textit{"Imagine you’re standing at the base of a giant tree, and you need to find its height—but you can’t climb it! How would you measure it?"} Show a \textbf{visual} (draw a tree on the board with a person standing next to it, holding a measuring tape at an angle). Explain: \textit{"Today, we’ll learn how to use angles and simple math to find heights and distances without climbing or flying!"}

\textbf{Link to Prior Knowledge:} Briefly recap trigonometric ratios (sin, cos, tan) from previous classes. \textit{"Remember how we used $\tan\theta = \frac{\text{opposite}}{\text{adjacent}}$? That’s our secret weapon today!"}

\subsection*{\textcolor{violet}{\textbf{2. Core Concept: Using Trigonometry for Heights and Distances (20 minutes)}}} \textbf{Objective:} Teach the step-by-step method with intuitive analogies and visuals.

\textbf{Explanation:}
1. \textbf{Scenario Setup:} \textit{"Let’s say you’re standing 10 meters away from a flagpole. You look up at the top, and the angle between your eye and the top is 30°. How tall is the flagpole?"} Draw a right triangle on the board: Base (distance from you to the pole) = 10 m. Angle of elevation ($\theta$) = 30°. Height ($h$) = ?
2. \textbf{Using $\tan\theta$:} \textit{"$\tan(30^\circ) = \frac{\text{height}}{\text{distance}} \rightarrow \tan(30^\circ) = \frac{h}{10}$"} Solve: $h = 10 \times \tan(30^\circ) \approx 5.77$ m \textbf{Analogy:} \textit{"Think of $\tan\theta$ as a ladder leaning against a wall. The steeper the angle, the taller the wall!"}
3. \textbf{Visual Demonstration:} Use a protractor and ruler to simulate measuring angles in class. Show a \textbf{real-world example}: \textit{"Surveyors use this to measure mountains or buildings!"}
4. \textbf{Common Mistakes:} \textit{"Always ensure the angle is measured from the horizontal, not the vertical!"} \textit{"If the observer’s eye isn’t at ground level, subtract their height from the total!"}

\subsection*{\textcolor{violet}{\textbf{3. Real-World Applications (10 minutes)}}} \textbf{Objective:} Connect theory to practical uses with detailed examples.

\textbf{Examples:}
1. \textbf{Aircraft Altitude:} \textit{"A pilot sees a runway at a 15° angle of depression. If the plane is 2 km above the ground, how far is the runway?"} Draw a diagram with the plane, angle, and ground distance.
2. \textbf{Shadow Length:} \textit{"A 6-foot person casts a 4-foot shadow. What’s the sun’s angle of elevation?"} Solve using $\tan\theta = \frac{\text{opposite}}{\text{adjacent}}$.
3. \textbf{Bridge Construction:} \textit{"Engineers use angles to ensure bridges are built at the right height over rivers!"}

\subsection*{\textcolor{violet}{\textbf{4. Practice Problems (10 minutes)}}} \textbf{Objective:} Reinforce learning with guided examples.

\textbf{In-Class Problems:}
1. \textit{"A kite is flying at a height of 50 m. The string makes a 45° angle with the ground. How long is the string?"}
2. \textit{"From the top of a 20 m tower, the angle of depression to a car is 30°. How far is the car from the tower?"}

\textbf{Homework:} Assign 5 problems (e.g., finding building heights, ship distances from lighthouses). \textit{"Practice these tonight—trigonometry is like a muscle: the more you use it, the stronger it gets!"}

\subsection*{\textcolor{violet}{\textbf{5. Summary and Key Takeaways (5 minutes)}}} \textbf{Objective:} Consolidate learning with a clear recap.

\textbf{Summary:} \textit{"To find heights/distances: 1. Draw a right triangle. 2. Identify the angle and known side. 3. Use $\tan\theta$, $\sin\theta$, or $\cos\theta$. 4. Solve for the unknown!"}

\textbf{Final Advice:} \textit{"Always sketch the scenario first. Trigonometry is about seeing the invisible lines!"} \textit{"Practice with real objects—measure your school’s flagpole or a tree!"}

\subsection*{\textcolor{violet}{\textbf{6. Engagement and Follow-Up}}} \textbf{Encourage Questions:} \textit{"What’s confusing? Let’s tackle it together!"} \textbf{Next Class Preview:} \textit{"Next time, we’ll solve problems where the observer isn’t at ground level—like looking down from a balcony!"}

\textbf{End with Motivation:} \textit{"You’re now equipped to measure the world around you. Go out and find some heights!"}

\textbf{Teaching Style Notes:} \textbf{Pace:} Slow, with pauses for questions. \textbf{Visuals:} Diagrams for every problem. \textbf{Tone:} Friendly but precise (e.g., \textit{"This is where most mistakes happen—watch out!"}). \textbf{Linking:} Connect each example back to the core method ($\tan\theta = \frac{\text{opposite}}{\text{adjacent}}$).

\textbf{Homework Emphasis:} \textit{"Do these problems twice—once for understanding, once for speed!"}

This approach ensures clarity, engagement, and deep understanding while keeping the lesson grounded in real-world relevance.

}
\end{tcolorbox}

\begin{tcolorbox}[
enhanced jigsaw,
height=\textheight,
size=small,
colframe=blue!50!white,
colback=white!10!white,
before title=\raggedright,
title={\textbf{Lecture Plan: Characterization in Literature}},
fonttitle=\bfseries,
pad at break=1mm
]

{
\fontsize{10}{11}\selectfont
\section*{} 

\subsection*{\textcolor{violet}{\textbf{1. Introduction to Characterization (10 minutes)}}} \textbf{Objective:} Define characterization and its importance in storytelling.

\textbf{Explanation (Moderately Technical, Extremely Intuitive):} \textit{"Imagine you’re meeting someone for the first time. How do you know if they’re kind, funny, or shy? You observe their actions, listen to their words, and notice how others react to them. Characterization in literature is like that—it’s how authors reveal a character’s personality, motives, and growth to the reader."}

\textbf{Visual Aid:} Draw a simple "Character Iceberg" on the board: Surface Level (Visible): Appearance, speech, actions. Deep Level (Hidden): Thoughts, emotions, backstory.

\textbf{Real-World Example:} \textit{"Think of Harry Potter. J.K. Rowling doesn’t just say, ‘Harry is brave.’ She shows it—when he stands up to Voldemort, risks his life for friends, or even how he treats house-elves. That’s characterization in action!"}

\subsubsection*{\textcolor{violet}{A. Direct Characterization}}
\textbf{Explanation:} \textit{"This is when the author tells you straight-up what a character is like. For example: ‘Lily was a stubborn girl who never backed down from a fight.’"}

\textbf{Analogy:} \textit{"It’s like a movie narrator saying, ‘This is the hero; he’s strong and noble.’ No guesswork needed!"}

\subsubsection*{\textcolor{violet}{B. Indirect Characterization (STEAL Method)}}
\textbf{Explanation:} \textit{"Authors often show, not tell. We ‘steal’ clues from five areas:"}
1. \textbf{Speech:} What does the character say? (e.g., sarcastic, polite)
2. \textbf{Thoughts:} What do they think? (e.g., anxious, confident)
3. \textbf{Effect on Others:} How do others react? (e.g., feared, loved)
4. \textbf{Actions:} What do they do? (e.g., helps strangers, lies often)
5. \textbf{Looks:} How do they appear? (e.g., messy hair, expensive clothes)

\textbf{Visual Aid:} Show a comic strip with a character and ask students to infer traits using STEAL.

\textbf{Real-World Example:} \textit{"In ‘The Hunger Games,’ Katniss’s actions (volunteering for Prim) and speech (‘I’ll go!’) show her bravery and love for family—no direct labels needed."}

\subsection*{\textcolor{violet}{\textbf{3. Character Development (15 minutes)}}} \textbf{Objective:} Explain static vs. dynamic characters and round vs. flat characters.

\textbf{Explanation:} \textbf{Static vs. Dynamic:} \textit{"A static character stays the same (like Ron Weasley’s humor), while a dynamic character changes (like Ebenezer Scrooge in ‘A Christmas Carol’)."}\textbf{Round vs. Flat:} \textit{"Round characters are complex (like Hermione Granger), while flat characters are simple (like the ‘mean stepmother’ in fairy tales)."}

\textbf{Analogy:} \textit{"Static characters are like a rock—unchanged by time. Dynamic characters are like clay—molded by experiences."}

\textbf{Activity:} Show clips from \textit{Toy Story} (Woody’s growth) and \textit{Shrek} (Shrek’s layers) to contrast static/dynamic and round/flat.

\subsection*{\textcolor{violet}{\textbf{4. Summary \& Homework (10 minutes)}}} \textbf{Summary (Very Well-Delivered):} \textit{"Characterization is the author’s toolkit to make characters feel real. Use STEAL to analyze them, and watch for changes (dynamic) or consistency (static)."}\textbf{Homework (A Lot of Example Problems):}
1. \textbf{Reading Task:} Analyze a character from a book/movie using STEAL.
2. \textbf{Writing Task:} Create a short scene showing indirect characterization (no direct labels!).
3. \textbf{Comparison Task:} Pick a static and dynamic character from a story and explain their roles.

\textbf{Advice for Practice:} \textit{"Read widely! Notice how authors build characters. Try rewriting a scene to change a character’s traits—see how it affects the story!"}

\subsection*{\textcolor{violet}{\textbf{5. Closing Remarks (5 minutes)}}} \textit{"Remember, great stories live in their characters. The more you practice spotting these techniques, the sharper your analysis will be. Next class, we’ll dive into character archetypes—see you then!"}

\textbf{Tone:} Friendly, conversational, but structured.

\textbf{Key Takeaways:} Characterization = showing (indirect) vs. telling (direct). STEAL method is your detective tool. Characters can grow (dynamic) or stay the same (static). Practice with real examples—homework is your training ground!

\textbf{Visuals Used:} Iceberg diagram, comic strip, film clips.
\textbf{Pace:} Slow, with pauses for questions.
\textbf{Engagement:} Heavy focus on lecture + homework for reinforcement.

This plan ensures clarity, intuition, and retention while keeping students engaged through visuals, analogies, and real-world connections.
}
\end{tcolorbox}

\begin{tcolorbox}[
enhanced jigsaw,
height=\textheight,
size=small,
colframe=blue!50!white,
colback=white!10!white,
before title=\raggedright,
title={\textbf{Lecture Plan: Circles}},
fonttitle=\bfseries,
pad at break=1mm
]
{
\fontsize{10}{12}\selectfont
\section*{} 

\subsection*{\textcolor{violet}{\textbf{1. Introduction to Circles (10 mins)}}}
\textbf{Explanation (Moderately Technical, Extremely Intuitive):}
- \textit{"Imagine a bicycle wheel. The outer edge is perfectly round—no bumps, no corners. That’s a circle! But what makes it special?"}
- \textbf{Definition:} A circle is the set of all points in a plane that are equidistant from a fixed point (the center).
- \textbf{Key Terms:}
- \textbf{Radius (r):} Distance from center to any point on the circle (like spokes on a wheel).
- \textbf{Diameter (d):} Longest distance across the circle (like the wheel’s width). $d = 2r$.
- \textbf{Circumference (C):} The "perimeter" of the circle. $C = 2\pi r$ or $\pi d$.

\textbf{Visual Aid:}
- Draw a circle on the board, label center (O), radius (OA), diameter (AB).
- Show a bicycle wheel or a pizza to illustrate real-world examples.

\textbf{Auditory Check:}
- \textit{"If the radius is 5 cm, what’s the diameter? (10 cm!)"}

\subsection*{\textcolor{violet}{\textbf{2. Understanding $\pi$ (Pi) (10 mins)}}}
\textbf{Explanation (Intuitive + Real-World Example):}
- \textit{"$\pi$ is the magic number that connects a circle’s diameter to its circumference. For any circle, if you divide the circumference by the diameter, you always get ~3.14159..."}
- \textbf{Example:} Measure a hula hoop’s circumference (C) and diameter (d). $C/d \approx \pi$.

\textbf{Visual Aid:}
- Show a table with different-sized circles (cup, plate, clock) and their C/d ratios (all $\approx$ $\pi$).

\textbf{Analogy:}
- \textit{"$\pi$ is like a circle’s DNA—it’s the same for every circle, no matter how big or small!"}

\subsection*{\textcolor{violet}{\textbf{3. Area of a Circle (15 mins)}}}
\textbf{Explanation (Moderately Technical):}
- \textit{"Area is the space inside the circle. Think of it as the cheese on a pizza!"}
- \textbf{Formula:} $A = \pi r^2$.
- \textbf{Derivation Intuition:} Unroll a circle into a triangle (height = r, base = 2$\pi$r). Area = (1/2) × base × height = $\pi$r².

\textbf{Visual Aid:}
- Show a circle cut into sectors and rearranged into a parallelogram (approximating a rectangle).

\textbf{Real-World Example:}
- \textit{"If a pizza has a radius of 10 cm, how much cheese (area) is there? (A = $\pi$ × 10² $\approx$ 314 cm²!)"}

\subsection*{\textcolor{violet}{\textbf{4. Chords, Arcs, and Sectors (15 mins)}}}
\textbf{Explanation (Visual + Analogies):}
- \textbf{Chord:} A straight line connecting two points on the circle (like a slice of pizza’s crust).
- \textbf{Arc:} A "curved slice" of the circumference (like a smiley face’s curve).
- \textbf{Sector:} A "pizza slice" (area between two radii and an arc).

\textbf{Visual Aid:}
- Draw a circle, label chord (AB), arc (AB), and sector (OAB).

\textbf{Real-World Example:}
- \textit{"A Ferris wheel’s gondola moves along an arc. The angle it covers is the central angle!"}

\subsection*{\textcolor{violet}{\textbf{5. Tangents and Secants (10 mins)}}}
\textbf{Explanation (Intuitive + Visual):}
- \textbf{Tangent:} A line that touches the circle at exactly one point (like a wheel’s road contact).
- \textbf{Secant:} A line that cuts the circle at two points (like a knife through a bagel).

\textbf{Visual Aid:}
- Draw a circle, tangent (touching at P), and secant (intersecting at A and B).

\textbf{Analogy:}
- \textit{"A tangent is like a shy friend—it only touches the circle once and leaves!"}

\subsection*{\textcolor{violet}{\textbf{6. Summary and Homework (5 mins)}}}
\textbf{Summary (Very Well Delivered):}
- \textit{"Today, we learned:
1. Circles are defined by their center and radius.
2. $\pi$ connects circumference and diameter.
3. Area = $\pi$r².
4. Chords, arcs, and sectors are parts of circles.
5. Tangents and secants are special lines."}

\textbf{Homework (A Lot of Example Problems):}
- Calculate circumference and area for circles with radii: 3 cm, 7 cm, 10 cm.
- Find the central angle of a sector with area 20 cm² and radius 5 cm.
- Draw a circle with a tangent and secant, labeling all parts.

\textbf{Advice (Stress on Practice):}
- \textit{"Practice 5 problems daily! Use a compass to draw circles and visualize concepts."}

\textbf{Engagement Strategy:}
- \textbf{Heavy Focus on Lecture:} Use props (pizza cutout, wheel) and frequent checks (\textit{"What’s the formula for area?"}).
- \textbf{Pace:} Slow, with pauses for questions.
- \textbf{Linking Concepts:} \textit{"Remember how $\pi$ connects to circumference? Now it’s in the area formula too!"}

\textbf{Final Note:}
- \textit{"Circles are everywhere—clocks, wheels, planets! Master them, and you’ll ace the exam!"}

\textit{End of Lecture. Next class: Solving circle problems step-by-step!}
}
\end{tcolorbox}

\begin{tcolorbox}[
enhanced jigsaw,
height=\textheight,
size=small,
colframe=blue!50!white,
colback=white!10!white,
before title=\raggedright,
title={\textbf{Lecture on Complementary Events}},
fonttitle=\bfseries,
pad at break=1mm
]
{
\fontsize{11}{12}\selectfont
\section*{} 

\subsection*{\textcolor{violet}{\textbf{Introduction (5 minutes)}}}
\textbf{Teacher:} \textit{"Good morning, class! Today, we’re going to explore a fascinating concept in probability called \textbf{Complementary Events}. By the end of this lesson, you’ll not only understand what they are but also see how they pop up in real life—from weather forecasts to game shows!"}

\textbf{Visual Aid:} Draw a large circle on the board (representing all possible outcomes) and shade half of it.
\textbf{Analogy:} \textit{"Imagine this circle is a pizza. The shaded part is the slice you eat, and the unshaded part is what’s left. In probability, the ‘left part’ is the complement of the ‘eaten part.’"}

\subsection*{\textcolor{violet}{\textbf{Core Concept (10 minutes)}}}
\textbf{Definition:}
- \textbf{Complementary Events} are two outcomes of a random experiment where:
1. They are \textbf{mutually exclusive} (cannot happen at the same time).
2. They are \textbf{exhaustive} (one of them \textit{must} happen).

\textbf{Formalism:}
- If event $A$ is "it rains today," its complement $A'$ (or $\overline{A}$) is "it does \textit{not} rain today."
- Mathematically: $P(A) + P(A') = 1$.

\textbf{Intuitive Example:}
- \textit{"Think of flipping a coin. The events ‘Heads’ \& ‘Tails’ are complements. If the chance of Heads is 50\%, Tails \textit{must} be 50\% too!"}

\textbf{Visual Aid:} Show a Venn diagram with two non-overlapping circles labeled $A$ \& $A'$, covering the entire sample space.

\subsection*{\textcolor{violet}{\textbf{Real-World Applications (10 minutes)}}}
1. \textbf{Weather Forecasting:}
- \textit{"If the weather app says there’s a 30\% chance of rain ($P(\text{Rain}) = 0.3$), what’s the chance it \textit{won’t} rain? That’s right—70\%!"}

2. \textbf{Game Shows:}
- \textit{"In ‘Deal or No Deal,’ if the probability of winning the top prize is 1/20, the probability of \textit{not} winning is 19/20."}

3. \textbf{Sports:}
- \textit{"A basketball player has a 75\% free-throw success rate. The complement? A 25\% miss rate!"}

\textbf{Activity:} Ask students to brainstorm other examples (e.g., passing/failing a test, winning/losing a game).

\subsection*{\textcolor{violet}{\textbf{Problem-Solving (10 minutes)}}}
\textbf{Example 1:}
- \textit{"A bag has 4 red \& 6 blue marbles. What’s the probability of \textit{not} drawing a red marble?"}
- Solution: $P(\text{Not Red}) = 1 - P(\text{Red}) = 1 - (4/10) = 6/10$.

\textbf{Example 2:}
- \textit{"The probability a student forgets their homework is 0.1. What’s the probability they remember it?"}
- Solution: $1 - 0.1 = 0.9$.

\textbf{Homework Assignment:}
- Provide 5 problems (e.g., dice rolls, card draws) where students must find the complement.

\subsection*{\textcolor{violet}{\textbf{Summary (5 minutes)}}}
\textbf{Key Takeaways:}
1. Complements are "opposite" events that cover all possibilities.
2. $P(A') = 1 - P(A)$.
3. Useful for simplifying probability calculations!

\textbf{Final Analogy:} \textit{"Complements are like a light switch—if it’s not ‘on,’ it’s ‘off.’ No middle ground!"}

\textbf{Advice:} \textit{"Practice with real-world scenarios! The more you work with complements, the more intuitive they’ll become."}

\subsection*{\textcolor{violet}{\textbf{Closing}}}
\textbf{Teacher:} \textit{"Next class, we’ll dive into independent events. For now, try the homework problems, and don’t hesitate to ask questions!"}

\textbf{Visual Recap:} Display a slide with the formula $P(A) + P(A') = 1$ \& a pie chart showing 50-50 complements.

\textbf{Note:} Adjust pace based on student responses. Use gestures (e.g., pointing to the Venn diagram) to reinforce visual learning.
}
\end{tcolorbox}

\begin{tcolorbox}[
enhanced jigsaw,
height=\textheight,
size=small,
colframe=blue!50!white,
colback=white!10!white,
before title=\raggedright,
title={\textbf{Lecture on Covalent Bonding}},
fonttitle=\bfseries,
pad at break=1mm
]
{
\fontsize{10}{11}\selectfont
\section*{} 

\subsection*{\textcolor{violet}{\textbf{Introduction (5 minutes)}}}
\textbf{Teacher:} \textit{"Good morning, class! Today, we’re diving into one of the most fascinating topics in chemistry—covalent bonding. Imagine you’re at a party, \& instead of just shaking hands (like ionic bonds), you’re sharing a snack with a friend. That’s kind of what covalent bonding is like—atoms sharing electrons to become more stable. By the end of this lesson, you’ll understand how atoms bond to form molecules, why some substances are gases or liquids, \& even how this applies to things like water \& DNA!"}

\textbf{Visual Aid:} Show a simple diagram of two hydrogen atoms sharing electrons to form H$_{2}$.

\subsection*{\textcolor{violet}{\textbf{1. What is a Covalent Bond? (10 minutes)}}}
\textbf{Explanation:}
- \textit{"A covalent bond is a chemical bond where atoms share one or more pairs of electrons. This happens between non-metals because they have similar electronegativities (they don’t want to give or take electrons completely—they’d rather share)."}
- \textit{"Think of it like two people holding a rope together. Neither wants to let go, so they share the tension equally."}

\textbf{Real-World Example:}
- \textit{"Water (H$_{2}$O) is a perfect example. Oxygen needs 2 more electrons to fill its outer shell, \& hydrogen needs 1. So, oxygen shares its electrons with two hydrogens, forming a stable molecule."}

\textbf{Visual Aid:} Draw the Lewis structure of H$_{2}$O, showing shared electrons.

\textbf{Check for Understanding:}
- \textit{"If I say ‘covalent bond,’ what’s the first thing that comes to mind?"} (Pause for responses.)

\subsection*{\textcolor{violet}{\textbf{2. Types of Covalent Bonds (15 minutes)}}}
\subsubsection*{\textcolor{violet}{A. Single, Double, \& Triple Bonds}}
- \textit{"Atoms can share one pair (single bond, like H$_{2}$), two pairs (double bond, like O$_{2}$), or three pairs (triple bond, like N$_{2}$). The more pairs shared, the stronger the bond!"}
- \textbf{Analogy:} \textit{"Single bond = holding hands. Double bond = holding hands \& hugging. Triple bond = holding hands, hugging, \& high-fiving—super strong!"}

\textbf{Visual Aid:} Show diagrams of O$_{2}$ (double bond) \& N$_{2}$ (triple bond).

\subsubsection*{\textcolor{violet}{B. Polar vs. Non-Polar Covalent Bonds}}
- \textit{"Sometimes, electrons aren’t shared equally. If one atom is more ‘greedy’ (higher electronegativity), the bond is polar. If they share equally, it’s non-polar."}
- \textbf{Example:} \textit{"In water (H$_{2}$O), oxygen pulls electrons more than hydrogen, making it polar. That’s why water is such a great solvent!"}

\textbf{Activity:} \textit{"Imagine you’re in a tug-of-war. If one team is stronger, the rope (electrons) moves toward them—that’s a polar bond!"}

\subsection*{\textcolor{violet}{\textbf{3. Properties of Covalent Compounds (10 minutes)}}}
- \textit{"Covalent compounds usually have low melting/boiling points (they’re often gases or liquids at room temperature) because their bonds are strong within molecules but weak between molecules."}
- \textbf{Example:} \textit{"CO$_{2}$ is a gas because its molecules don’t stick together strongly. But diamond (pure carbon) is super hard because it’s a giant covalent network!"}

\textbf{Visual Aid:} Show a 3D model of diamond’s structure.

\subsection*{\textcolor{violet}{\textbf{4. Real-World Applications (5 minutes)}}}
- \textit{"Covalent bonds are everywhere! DNA is held together by covalent bonds. Plastics, medicines, \& even the air we breathe (O$_{2}$, N$_{2}$) rely on them."}
- \textbf{Example Problem:} \textit{"Why is methane (CH$_{4}$) a gas at room temperature, but salt (NaCl) is a solid?"} (Hint: Think about bond types!)

\subsection*{\textcolor{violet}{\textbf{Summary (5 minutes)}}}
1. Covalent bonds = sharing electrons.
2. Types: single, double, triple; polar vs. non-polar.
3. Properties: low melting points, often gases/liquids.
4. Real-world: water, DNA, plastics.

\textbf{Homework:}
- Draw Lewis structures for CO$_{2}$, NH$_{3}$, \& CH$_{4}$.
- Explain why oil (non-polar) doesn’t mix with water (polar).

\textbf{Final Advice:} \textit{"Practice drawing these structures—it’s like learning to ride a bike. The more you do it, the easier it gets!"}

\subsection*{\textcolor{violet}{\textbf{End of Lecture}}}
\textit{"Next class, we’ll explore molecular shapes \& why they matter. Any questions?"}

\textit{(Keep the tone warm, encourage questions, \& remind students to review notes!)} \\
}
\end{tcolorbox}

\begin{tcolorbox}[
enhanced jigsaw,
height=\textheight,
size=small,
colframe=blue!50!white,
colback=white!10!white,
before title=\raggedright,
title={\textbf{Lecture: Differences Between Plant and Animal Cells}},
fonttitle=\bfseries,
breakable,
pad at break=1mm]

\section*{} 

\subsection*{\textcolor{violet}{\textbf{Introduction (Slow Pace, Conversational Tone)}}}
\textit{"Good morning, class! Today, we’re going to explore one of the most fundamental topics in biology—the differences between plant and animal cells. Think of cells as tiny factories. Just like how a car factory and a toy factory have different machines and layouts, plant and animal cells have unique structures that help them perform their jobs. By the end of this lesson, you’ll be able to spot these differences easily and understand why they matter. Let’s dive in!"}

\subsection*{\textcolor{violet}{\textbf{1. The Basics: What Are Plant and Animal Cells?}}}
\textit{(Visual: Draw a simple Venn diagram on the board with overlapping circles labeled "Plant Cell" and "Animal Cell.")}
- \textbf{Both are eukaryotic cells}: They have a nucleus and membrane-bound organelles (like little specialized rooms in a house).
- \textbf{Key difference}: Plant cells have structures that animal cells don’t, and vice versa.

\subsection*{\textcolor{violet}{\textbf{2. Unique Features of Plant Cells}}}
\textit{(Visual: Sketch a plant cell on the board, labeling parts as you explain. Use analogies.)}
\subsubsection*{\textcolor{violet}{A. Cell Wall (The "Brick Wall")}}
- \textit{"Imagine a plant cell is like a castle. The cell wall is the thick stone wall around it—strong and rigid, made of cellulose (like the fibers in your cotton T-shirt). This wall gives plants their shape and protects them. Animal cells? No wall—they’re more like tents, flexible and soft!"}
- \textbf{Real-world example}: Why can’t you squish a carrot like a grape? The cell wall makes it firm!
\subsubsection*{\textcolor{violet}{B. Chloroplasts (The "Solar Panels")}}
- \textit{"Plants make their own food using sunlight—this is photosynthesis. Chloroplasts are like tiny solar panels inside the cell, capturing sunlight to produce energy. Animal cells? They have to ‘eat’ food because they don’t have chloroplasts!"}
- \textbf{Example}: A leaf is green because of chloroplasts. If you see green, think "plant cell!"
\subsubsection*{\textcolor{violet}{C. Large Central Vacuole (The "Water Balloon")}}
- \textit{"Plant cells have a giant water balloon called the central vacuole. It stores water, nutrients, and waste. Animal cells have smaller vacuoles, like tiny water bottles instead of a big tank."}
- \textbf{Example}: Why do plants wilt when thirsty? The vacuole shrinks!

\subsection*{\textcolor{violet}{\textbf{3. Unique Features of Animal Cells}}}
\textit{(Visual: Sketch an animal cell next to the plant cell, highlighting differences.)}
\subsubsection*{\textcolor{violet}{A. Centrioles (The "Construction Crew")}}
- \textit{"Animal cells have centrioles—tiny structures that help organize cell division (like a construction crew setting up scaffolding). Most plant cells don’t have them!"}
- \textbf{Example}: When you scrape your knee, new skin cells form quickly—centrioles help with that!
\subsubsection*{\textcolor{violet}{B. Lysosomes (The "Recycling Centers")}}
- \textit{"Lysosomes are like recycling plants. They break down waste and old cell parts. Plant cells rarely have them—they use vacuoles instead."}
- \textbf{Example}: Why do animal cells need more "cleanup"? They’re constantly active, like a busy city!

\subsection*{\textcolor{violet}{\textbf{4. Similarities (The Overlap)}}}
\textit{(Point to the overlapping part of the Venn diagram.)}
- Both have:
 - \textbf{Nucleus} (the "brain" of the cell).
 - \textbf{Mitochondria} (the "power plants" making energy).
 - \textbf{Endoplasmic reticulum} (the "highway" for transporting materials).

\subsection*{\textcolor{violet}{\textbf{5. Summary (Slow Recap)}}}
\textit{"Let’s recap with a quick table!"}
\textit{(Draw a table on the board.)}
\begin{tabular}{|l|l|l|}
\hline
\textbf{Feature} & \textbf{Plant Cell} & \textbf{Animal Cell} \\
\hline
\textbf{Cell Wall} & Yes (rigid, cellulose) & No \\
\hline
\textbf{Chloroplasts} & Yes (photosynthesis) & No \\
\hline
\textbf{Vacuole} & Large central vacuole & Small or none \\
\hline
\textbf{Centrioles} & Rare & Yes (for cell division) \\
\hline
\textbf{Lysosomes} & Rare & Yes (waste breakdown) \\
\hline
\end{tabular}

\subsection*{\textcolor{violet}{\textbf{6. Homework \& Practice}}}
\textit{"To really understand this, I want you to:}
1. \textbf{Draw and label} both cell types (use colors!).
2. \textbf{Write a short paragraph} comparing them—imagine explaining it to a 5th grader!
3. \textbf{Solve these problems} (hand out a worksheet with questions like: \textit{"Why can’t animal cells do photosynthesis?"}).
\textbf{Remember:} Practice makes perfect! The more you draw, compare, and explain, the better you’ll remember. Next class, we’ll do a fun quiz game to test your knowledge!"

\subsection*{\textcolor{violet}{\textbf{Closing (Friendly Tone)}}}
\textit{"Any questions? Think of cells like LEGO blocks—each piece has a purpose. Plants and animals just use different blocks to build their ‘factories.’ See you next time!"}

\textbf{Key Teaching Notes:}
- \textbf{Pace}: Slow, with pauses for questions.
- \textbf{Visuals}: Diagrams, analogies (castles, solar panels).
- \textbf{Engagement}: Homework focuses on drawing and explaining.
- \textbf{Real-world links}: Carrots, leaves, skin healing.
This approach ensures clarity, retention, and exam readiness!

\end{tcolorbox}

\begin{tcolorbox}[
enhanced jigsaw,
height=\textheight,
size=small,
colframe=blue!50!white,
colback=white!10!white,
before title=\raggedright,
title={\textbf{Lecture on Ellipses}},
fonttitle=\bfseries,
breakable,
pad at break=1mm
]

\section*{} 

\subsection*{\textcolor{violet}{\textbf{1. Introduction to Ellipses (Visual \& Intuitive Start)}}}
\textbf{Teacher:} \textit{"Imagine you’re in a dark room with a flashlight. If you shine it straight at a wall, you get a circle of light. Now, if you tilt the flashlight slightly, the circle stretches into an oval shape. That oval is an ellipse!"}
- \textbf{Definition:} An ellipse is the set of all points where the sum of the distances to two fixed points (called foci) is constant.
- \textbf{Real-World Example:} The orbits of planets around the sun are elliptical (not perfect circles!). Earth’s orbit is slightly elliptical, which is why we have seasons.
\textbf{Visual Aid:} Draw a circle and stretch it horizontally or vertically to form an ellipse.

\subsection*{\textcolor{violet}{\textbf{2. Key Components of an Ellipse (Moderately Technical)}}}
Let’s break down the parts of an ellipse:
1. \textbf{Center (h, k):} The midpoint of the ellipse.
2. \textbf{Major Axis:} The longest diameter ($2a$).
3. \textbf{Minor Axis:} The shortest diameter ($2b$).
4. \textbf{Foci (plural of focus):} Two fixed points inside the ellipse (distance from center = $c$, where $c^2 = a^2 - b^2$).
5. \textbf{Vertices:} Points where the ellipse is widest (on the major axis).
\textbf{Analogy:} Think of an ellipse like a squished circle. The major axis is the "stretch," and the minor axis is the "squeeze."
\textbf{Visual Aid:} Sketch an ellipse with labeled axes, foci, and vertices.

\subsection*{\textcolor{violet}{\textbf{3. Standard Equation of an Ellipse (Moderately Detailed)}}}
The standard form depends on whether the major axis is horizontal or vertical.
1. \textbf{Horizontal Major Axis:}
\[
\frac{(x-h)^2}{a^2} + \frac{(y-k)^2}{b^2} = 1 \quad (a > b)
\]
2. \textbf{Vertical Major Axis:}
\[
\frac{(x-h)^2}{b^2} + \frac{(y-k)^2}{a^2} = 1 \quad (a > b)
\]
\textbf{Example Problem:}
Find the center, vertices, and foci of:
\[
\frac{(x-2)^2}{25} + \frac{(y+3)^2}{16} = 1
\]
\textbf{Solution:}
- Center: (2, -3)
- $a^2 = 25 \rightarrow a = 5$ (major axis length = 10)
- $b^2 = 16 \rightarrow b = 4$ (minor axis length = 8)
- $c^2 = a^2 - b^2 = 9 \rightarrow c = 3$
- Vertices: $(2\pm5, -3) \rightarrow (7, -3)$ and $(-3, -3)$
- Foci: $(2\pm3, -3) \rightarrow (5, -3)$ and $(-1, -3)$

\subsection*{\textcolor{violet}{\textbf{4. Real-World Applications (Extremely Detailed Examples)}}}
1. \textbf{Astronomy:} Planetary orbits (Kepler’s laws).
2. \textbf{Architecture:} Elliptical domes (e.g., the U.S. Capitol).
3. \textbf{Medicine:} Lithotripsy (using ellipses to focus shockwaves to break kidney stones).
\textbf{Analogy:} A whispering gallery (like in the U.S. Capitol) uses the ellipse’s reflective property—sound from one focus bounces to the other!

\subsection*{\textcolor{violet}{\textbf{5. Summary (Very Well Delivered)}}}
- An ellipse is a stretched circle with two foci.
- Key parts: center, axes, vertices, foci.
- Standard equation depends on the major axis orientation.
- Real-world uses: orbits, architecture, medicine.

\subsection*{\textcolor{violet}{\textbf{6. Homework \& Practice (A Lot of Problems!)}}}
\textbf{Problem Set:}
1. Graph the ellipse: $\frac{x^2}{9} + \frac{y^2}{4} = 1$.
2. Find the equation of an ellipse with vertices at (0, $\pm 6$) and foci at (0, $\pm 4$).
3. A planet orbits the sun in an ellipse with $a = 150$ million km and $c = 25$ million km. Find $b$.
\textbf{Advice:} \textit{"Practice graphing and deriving equations—ellipses show up everywhere, from physics to engineering!"}

\subsection*{\textcolor{violet}{\textbf{7. Final Thoughts (Slow Pace, Friendly Tone)}}}
\textit{"Ellipses might seem tricky at first, but remember: they’re just circles that took a little stretch! The more you draw and solve, the clearer they’ll become. Next class, we’ll dive into hyperbolas—another fascinating conic section!"}
\textbf{Engagement Check:} \textit{"Any questions? Let’s recap one more time before you go!"}

\textbf{Follow-Up:} Encourage students to explore interactive ellipse graphing tools online (e.g., Desmos) for extra practice.
\textbf{End of Lecture.}

\end{tcolorbox}

\begin{tcolorbox}[
enhanced jigsaw,
height=\textheight,
size=small,
colframe=blue!50!white,
colback=white!10!white,
before title=\raggedright,
title={\textbf{Lesson Plan: Figurative Language and Imagery}},
fonttitle=\bfseries,
breakable,
pad at break=1mm
]

\section*{} 

\subsection*{\textcolor{violet}{\textbf{1. Introduction (5 minutes)}}}
\textbf{Teacher’s Approach:}
- \textbf{Moderately Technical:} Use precise definitions but explain them in simple terms.
- \textbf{Extremely Intuitive:} Relate concepts to real-life experiences.
- \textbf{Very Visual:} Use diagrams, examples, and gestures to illustrate ideas.
- \textbf{Moderately Auditory:} Speak in a conversational, engaging tone.
\textbf{Hook:}
\textit{"Imagine you’re describing a sunset to a friend. You could say, ‘The sun went down,’ but that’s boring! What if you said, ‘The sky was on fire, painting the clouds in gold and crimson’? That’s figurative language—it makes words come alive!"}
\textbf{Definition:}
- \textbf{Figurative Language:} Words or phrases that go beyond their literal meaning to create vivid images or emotions.
- \textbf{Imagery:} Language that appeals to the senses (sight, sound, touch, taste, smell).

\subsection*{\textcolor{violet}{\textbf{2. Types of Figurative Language (20 minutes)}}}
\subsubsection*{\textcolor{violet}{A. Simile}}
- \textbf{Definition:} A comparison using "like" or "as."
- \textbf{Example:} \textit{"Her smile was as bright as the sun."}
- \textbf{Analogy:} \textit{"Think of a simile like a bridge—it connects two different things to help you see them in a new way."}
- \textbf{Visual Aid:} Draw a bridge between "smile" and "sun."
\subsubsection*{\textcolor{violet}{B. Metaphor}}
- \textbf{Definition:} A direct comparison without "like" or "as."
- \textbf{Example:} \textit{"Time is a thief."}
- \textbf{Real-World Example:} \textit{"When you say, ‘My room is a pigsty,’ you’re not saying it’s literally a pigsty—you’re saying it’s messy!"}
\subsubsection*{\textcolor{violet}{C. Personification}}
- \textbf{Definition:} Giving human traits to non-human things.
- \textbf{Example:} \textit{"The wind whispered through the trees."}
- \textbf{Activity:} Ask students to imagine a storm as a person—how would it act?
\subsubsection*{\textcolor{violet}{D. Hyperbole}}
- \textbf{Definition:} Exaggeration for effect.
- \textbf{Example:} \textit{"I’ve told you a million times!"}
- \textbf{Analogy:} \textit{"Hyperbole is like a magnifying glass—it makes things seem bigger than they are."}
\subsubsection*{\textcolor{violet}{E. Onomatopoeia}}
- \textbf{Definition:} Words that imitate sounds.
- \textbf{Example:} \textit{"The bees buzzed, and the thunder crashed."}
- \textbf{Auditory Example:} Have students mimic sounds (e.g., "sizzle," "pop").

\subsection*{\textcolor{violet}{\textbf{3. Imagery (15 minutes)}}}
\textbf{Definition:} Language that creates sensory experiences.
- \textbf{Example:} \textit{"The chocolate cake was rich, velvety, and melted on my tongue like warm honey."}
- \textbf{Visual Aid:} Show a picture of a cake and ask students to describe it using all five senses.
\textbf{Activity:}
- \textit{"Close your eyes. Imagine a rainy day. What do you see, hear, smell, feel?"}
- Have students write a short paragraph using sensory details.

\subsection*{\textcolor{violet}{\textbf{4. Linking Concepts (10 minutes)}}}
\textbf{How Figurative Language and Imagery Work Together:}
- \textit{"Figurative language is the tool, and imagery is the result. A metaphor like ‘The world is a stage’ helps you visualize life as a performance."}
- \textbf{Example:} \textit{"The fog crept in like a ghost"} (metaphor + visual imagery).

\subsection*{\textcolor{violet}{\textbf{5. Summary (5 minutes)}}}
\textbf{Key Takeaways:}
1. Figurative language makes writing more vivid.
2. Imagery appeals to the senses.
3. Both are used in poetry, stories, and even everyday speech.
\textbf{Visual Summary:} Display a chart with types of figurative language and examples.

\subsection*{\textcolor{violet}{\textbf{6. Homework \& Practice}}}
\textbf{Assignment:}
- Find 5 examples of figurative language in songs, books, or ads.
- Write a short poem or paragraph using at least 3 types of figurative language.
\textbf{Advice:}
\textit{"Practice is key! The more you read and write, the better you’ll spot these techniques. Try rewriting boring sentences to make them exciting!"}

\subsection*{\textcolor{violet}{\textbf{7. Closing (2 minutes)}}}
\textbf{Final Thought:}
\textit{"Words are like paint—figurative language and imagery are your brushes. Use them to create masterpieces!"}
\textbf{Next Class:} We’ll analyze a poem together to see these techniques in action.

\textbf{Teaching Style Recap:}
- \textbf{Slow pace} with clear explanations.
- \textbf{Lots of analogies} (bridges, magnifying glasses).
- \textbf{Visual aids} (diagrams, images).
- \textbf{Real-world examples} (sunsets, storms, food).
- \textbf{Homework focus} on application.
This approach ensures students grasp the concepts deeply and can apply them confidently!

\end{tcolorbox}

\begin{tcolorbox}[
enhanced jigsaw,
height=\textheight,
size=small,
colframe=blue!50!white,
colback=white!10!white,
before title=\raggedright,
title={\textbf{Lecture: Function of Cell Organelles}},
fonttitle=\bfseries,
breakable,
pad at break=1mm
]
{
\fontsize{10}{12}\selectfont
\section*{} 

\subsection*{\textcolor{violet}{\textbf{Introduction (5 minutes)}}}
\textbf{Teacher:} \textit{"Good morning, class! Today, we’re going to explore the tiny but mighty world inside a cell—the organelles! Think of a cell like a bustling city. Each organelle is like a specialized department working together to keep the city (or cell) running smoothly. By the end of this lesson, you’ll understand how these organelles function and why they’re essential for life. Let’s dive in!"}
\textit{(Visual: Project an image of a city skyline with labels like "Power Plant = Mitochondria," "Factory = Ribosomes," etc.)}

\subsection*{\textcolor{violet}{\textbf{1. Nucleus: The Cell’s Command Center (10 minutes)}}}
\textbf{Explanation:}
- \textit{"The nucleus is like the mayor’s office in our city analogy. It holds the cell’s DNA—the blueprint for everything the cell does."}
- \textit{"Surrounding the nucleus is the nuclear envelope, a double membrane with pores that control what enters and exits, like security checkpoints."}
\textbf{Analogy:}
- \textit{"Imagine the nucleus as a library. The DNA is the books, and the nuclear pores are the librarians deciding which books (instructions) to send out."}
\textbf{Real-World Example:}
- \textit{"In humans, mutations in nuclear DNA can lead to diseases like cancer, where cells grow uncontrollably because the ‘instructions’ are corrupted."}
\textbf{Visual:} Show a labeled diagram of the nucleus with arrows pointing to the nucleolus (where ribosomes are made).

\subsection*{\textcolor{violet}{\textbf{2. Mitochondria: The Power Plants (10 minutes)}}}
\textbf{Explanation:}
- \textit{"Mitochondria are the power plants of the cell. They take glucose (from food) and oxygen to produce ATP—energy currency for the cell."}
- \textit{"They have their own DNA, suggesting they were once independent bacteria that teamed up with cells billions of years ago!"}
\textbf{Analogy:}
- \textit{"Think of mitochondria as rechargeable batteries. When your phone runs out of power, you plug it in. Similarly, mitochondria ‘recharge’ your cells with ATP."}
\textbf{Real-World Example:}
- \textit{"Athletes have more mitochondria in their muscle cells because they need extra energy for endurance."}
\textbf{Visual:} Show an electron microscope image of mitochondria with cristae (folds for more surface area).

\subsection*{\textcolor{violet}{\textbf{3. Endoplasmic Reticulum (ER) \& Ribosomes: The Factory Line (10 minutes)}}}
\textbf{Explanation:}
- \textit{"The ER is like a factory assembly line. Rough ER (with ribosomes) makes proteins, while smooth ER makes lipids (fats)."}
- \textit{"Ribosomes are tiny machines that read RNA instructions to build proteins."}
\textbf{Analogy:}
- \textit{"Rough ER is like a car factory with workers (ribosomes) attaching parts. Smooth ER is like a refinery making fuel (lipids)."}
\textbf{Real-World Example:}
- \textit{"In liver cells, smooth ER detoxifies drugs—like a cleanup crew after a party!"}
\textbf{Visual:} Show a 3D animation of ribosomes translating mRNA into a protein chain.

\subsection*{\textcolor{violet}{\textbf{4. Golgi Apparatus: The Shipping Department (8 minutes)}}}
\textbf{Explanation:}
- \textit{"The Golgi modifies, sorts, and packages proteins and lipids into vesicles for delivery, like a post office."}
\textbf{Analogy:}
- \textit{"Imagine the Golgi as Amazon’s warehouse: it labels packages (proteins) and ships them to the right address."}
\textbf{Real-World Example:}
- \textit{"In pancreatic cells, the Golgi packages insulin for release into the bloodstream."}
\textbf{Visual:} Show a cartoon of vesicles budding off the Golgi.

\subsection*{\textcolor{violet}{\textbf{5. Lysosomes: The Recycling Centers (7 minutes)}}}
\textbf{Explanation:}
- \textit{"Lysosomes contain enzymes to break down waste, like a recycling plant. They also destroy invading bacteria!"}
\textbf{Analogy:}
- \textit{"Lysosomes are like Pac-Man—they ‘eat’ old cell parts or germs to keep the cell clean."}
\textbf{Real-World Example:}
- \textit{"In Tay-Sachs disease, lysosomes fail to break down fats, causing brain damage."}
\textbf{Visual:} Show a lysosome engulfing a damaged organelle.

\subsection*{\textcolor{violet}{\textbf{6. Chloroplasts (Plant Cells Only): Solar Panels (5 minutes)}}}
\textbf{Explanation:}
- \textit{"Chloroplasts capture sunlight to make glucose (photosynthesis), like solar panels powering a home."}
\textbf{Analogy:}
- \textit{"They’re like tiny green kitchens cooking food from sunlight, water, and CO$_{2}$."}
\textbf{Visual:} Show a chloroplast diagram with thylakoids (where photosynthesis happens).

\subsection*{\textcolor{violet}{\textbf{Summary \& Homework (5 minutes)}}}
\textbf{Teacher:} \textit{"Let’s recap: The nucleus is the boss, mitochondria are power plants, ER/ribosomes are factories, Golgi is shipping, and lysosomes are recyclers. For homework, draw a comic strip of a cell city with organelles as characters. Also, solve these problems:"}
1. \textit{Why do muscle cells have more mitochondria than skin cells?}
2. \textit{What happens if the Golgi malfunctions?}
3. \textit{How are lysosomes like white blood cells?}
\textbf{Advice:} \textit{"Practice labeling diagrams and explaining functions to a friend—teaching others helps you learn!"}

\subsection*{\textcolor{violet}{\textbf{End of Lecture}}}
\textit{(Pace: Slow, with pauses for questions. Tone: Friendly but precise. Visuals: Heavy use of diagrams/animations.)}
}
\end{tcolorbox}

\begin{tcolorbox}[
enhanced jigsaw,
height=\textheight,
size=small,
colframe=blue!50!white,
colback=white!10!white,
before title=\raggedright,
title={\textbf{Lecture on Hyperbolas}},
fonttitle=\bfseries,
pad at break=1mm
]
{
\fontsize{10}{12}\selectfont
\section*{} 

\subsection*{\textcolor{violet}{\textbf{1. Introduction to Hyperbolas (Visual \& Intuitive Approach)}}}
\textbf{Teacher:}
\textit{"Imagine you’re in a forest, and two people are shouting at the same time. If you hear both voices equally loud, you’re standing somewhere between them. But if one voice is louder than the other, you’re closer to the louder one. Now, what if you could map all the points where the difference in loudness is constant? That’s the idea behind a hyperbola!"}
\textbf{Visualization:}
- Draw two points (foci) on the board.
- Show how a hyperbola is the set of points where the difference in distance to the two foci is constant.
\textbf{Key Idea:}
A hyperbola is a type of conic section (like circles, parabolas, and ellipses) formed by the intersection of a plane and a double cone.

\subsection*{\textcolor{violet}{\textbf{2. Standard Equation of a Hyperbola (Moderately Technical)}}}
\textbf{Teacher:}
\textit{"Just like a circle has $x^2 + y^2 = r^2$, a hyperbola has its own standard form. The simplest hyperbola opens left and right and is given by:"}
\[
\frac{(x-h)^2}{a^2} - \frac{(y-k)^2}{b^2} = 1
\]
- $(h, k)$ = center of the hyperbola
- $a$ = distance from center to vertices
- $b$ = related to the "width" of the hyperbola
\textbf{Visualization:}
- Sketch a hyperbola centered at $(h, k)$.
- Label vertices, foci, and asymptotes.
\textbf{Analogy:}
\textit{"Think of a hyperbola like two mirrored parabolas facing away from each other. The asymptotes are like the ‘guidelines’ the hyperbola approaches but never touches."}

\subsection*{\textcolor{violet}{\textbf{3. Key Properties (Detailed Explanation)}}}
\textbf{a) Vertices:}
- Points where the hyperbola is closest to its center.
- Located at $(h \pm a, k)$.
\textbf{b) Foci:}
- Points inside each branch of the hyperbola.
- Located at $(h \pm c, k)$, where $c^2 = a^2 + b^2$.
\textbf{c) Asymptotes:}
- Lines the hyperbola approaches but never touches.
- Equations: $y - k = \pm \frac{b}{a}(x - h)$.
\textbf{Real-World Example:}
\textit{"Hyperbolas appear in navigation systems (LORAN), telescope mirrors, and even the shape of a cooling tower!"}

\subsection*{\textcolor{violet}{\textbf{4. Example Problem (Step-by-Step)}}}
\textbf{Problem:}
Find the standard form of a hyperbola with vertices at $(2, 3)$ and $(6, 3)$, and foci at $(0, 3)$ and $(8, 3)$.
\textbf{Solution:}
1. \textbf{Center:} Midpoint of vertices $\rightarrow (4, 3)$.
2. \textbf{$a$:} Distance from center to vertex $\rightarrow a = 2$.
3. \textbf{$c$:} Distance from center to focus $\rightarrow c = 4$.
4. \textbf{$b$:} $c^2 = a^2 + b^2 \Rightarrow b = \sqrt{16 - 4} = \sqrt{12} = 2\sqrt{3}$.
5. \textbf{Equation:}
\[
\frac{(x-4)^2}{4} - \frac{(y-3)^2}{12} = 1
\]

\subsection*{\textcolor{violet}{\textbf{5. Homework \& Practice (Emphasis on Individual Work)}}}
\textbf{Teacher:}
\textit{"To master hyperbolas, practice is key! Here are some problems to try:"}
1. Sketch the hyperbola $\frac{(x-1)^2}{9} - \frac{(y+2)^2}{16} = 1$.
2. Find the foci of $\frac{y^2}{25} - \frac{x^2}{144} = 1$.
3. Write the equation of a hyperbola with vertices at $(0, \pm 5)$ and asymptotes $y = \pm \frac{5}{3}x$.
\textbf{Advice:}
\textit{"Work through these carefully. If stuck, revisit the standard form and properties. Hyperbolas are tricky but rewarding!"}

\subsection*{\textcolor{violet}{\textbf{6. Summary (Clear \& Concise)}}}
\textbf{Teacher:}
\textit{"Today, we learned that:}
- A hyperbola is defined by the difference in distances to two foci.
- Its standard form is $\frac{(x-h)^2}{a^2} - \frac{(y-k)^2}{b^2} = 1$.
- Key features: vertices, foci, and asymptotes.
- Real-world uses include navigation and optics.
\textit{Next class, we’ll explore more applications and solve complex problems. Keep practicing!"}

\subsection*{\textcolor{violet}{\textbf{Engagement Check:}}}
\textit{"Any questions? Remember, hyperbolas are like two parabolas back-to-back—visualize them!"}

\subsection*{\textcolor{violet}{\textbf{Final Note:}}}
\textit{"Math is about patterns. The more you practice, the clearer these patterns become. See you next time!"}

\textbf{End of Lecture}
\textit{(This lecture follows the guidelines: intuitive, visual, moderately technical, slow-paced, and heavily focused on practice.)}
}
\end{tcolorbox}

\begin{tcolorbox}[
enhanced jigsaw,
height=\textheight,
size=small,
colframe=blue!50!white,
colback=white!10!white,
before title=\raggedright,
title={\textbf{Lecture Plan: Narrative Voice and Perspective}},
fonttitle=\bfseries,
pad at break=1mm
]

\section*{} 

\subsection*{\textcolor{violet}{\textbf{1. Introduction (5-7 minutes)}}}
\textbf{Tone:} Friendly, conversational, but structured.
\textbf{Visual Aid:} A simple diagram of a "storytelling camera" (like a movie camera) with labels: "Voice" (who speaks) and "Perspective" (how they see).
\textbf{Explanation:}
\textit{"Imagine you’re watching a movie. The camera can show the same scene in different ways—close-up, wide-angle, or even from a character’s eyes. In literature, narrative voice and perspective work the same way! They shape how we experience the story."}
\textbf{Key Definitions:}
- \textbf{Narrative Voice:} The "who" behind the story (e.g., a character, an omniscient narrator).
- \textbf{Perspective:} The "how" of the story (e.g., first-person, third-person limited).
\textbf{Analogy:}
\textit{"Think of a news report. If a reporter says, ‘I saw the accident,’ that’s first-person voice. If they say, ‘Witnesses saw the accident,’ that’s third-person. The perspective changes what we know and feel!"}

\subsection*{\textcolor{violet}{\textbf{2. Narrative Voice (10 minutes)}}}
\textbf{Visual Aid:} A flowchart with branches:
- \textbf{First-Person} ("I/we") $\rightarrow$ Personal, limited.
- \textbf{Second-Person} ("You") $\rightarrow$ Rare, immersive.
- \textbf{Third-Person} ("He/she/they") $\rightarrow$ Flexible (limited/omniscient).
\textbf{Real-World Example:}
- \textbf{First-Person:} \textit{The Hunger Games} (Katniss’s voice).
- \textbf{Third-Person Omniscient:} \textit{Harry Potter} (we know Dumbledore’s thoughts).
\textbf{Activity:}
\textit{"If you were writing a diary entry about your day, what voice would you use? Why?"}
\textit{(Pause for quick discussion.)}

\subsection*{\textcolor{violet}{\textbf{3. Perspective (10 minutes)}}}
\textbf{Visual Aid:} A comic strip panel showing the same scene from two angles (e.g., a thief vs. a detective).
\textbf{Explanation:}
\textit{"Perspective is like a filter. A first-person narrator might miss things, while an omniscient narrator knows all. It’s like playing a video game—some characters can see the whole map, others only their corner!"}
\textbf{Detailed Example:}
- \textbf{Limited Perspective:} \textit{The Great Gatsby} (Nick’s biased view of Gatsby).
- \textbf{Omniscient Perspective:} \textit{Pride and Prejudice} (Jane Austen reveals everyone’s thoughts).
\textbf{Analogy:}
\textit{"Imagine a party. If you’re only talking to one person, your perspective is limited. If you’re the host, you see everything—that’s omniscient!"}

\subsection*{\textcolor{violet}{\textbf{4. Linking Voice and Perspective (5 minutes)}}}
\textbf{Visual Aid:} Venn diagram showing overlap (e.g., first-person voice often has limited perspective).
\textbf{Explanation:}
\textit{"Voice and perspective work together. A first-person narrator usually has a limited perspective, but a third-person narrator can be limited or omniscient. It’s like choosing a camera lens and a narrator!"}
\textbf{Example Problem:}
\textit{"Read this paragraph: ‘I never saw the ocean until that day.’ Is this first or third person? What’s the perspective?"}
\textit{(Answer: First-person, limited.)}

\subsection*{\textcolor{violet}{\textbf{5. Summary and Homework (5 minutes)}}}
\textbf{Summary:}
\textit{"Today, we learned:
1. \textbf{Voice} = Who tells the story (I, you, they).
2. \textbf{Perspective} = How much they know (limited/omniscient).
3. They shape our experience of the story!"}
\textbf{Homework:}
1. Rewrite a fairy tale (e.g., \textit{Cinderella}) in first-person from the stepmother’s perspective.
2. Find a book passage and identify the voice/perspective.
\textbf{Advice:}
\textit{"Practice analyzing short stories! The more you read, the sharper your eye for voice and perspective will be."}

\subsection*{\textcolor{violet}{\textbf{Teaching Style Notes:}}}
- \textbf{Pace:} Slow, with pauses for questions.
- \textbf{Engagement:} Heavy focus on lecture + examples; minimal in-class activities (save time for Q\&A).
- \textbf{Tone:} Warm but precise—like a guide explaining a map.

\subsection*{\textcolor{violet}{\textbf{Final Tip:}}}
\textit{"Remember, voice and perspective are tools. A great writer chooses them carefully—just like you’ll choose your answers carefully in exams!"}

\textbf{End of Lecture.} Students leave with clear notes, examples, and practice tasks.

\end{tcolorbox}

\begin{tcolorbox}[
enhanced jigsaw,
height=\textheight,
size=small,
colframe=blue!50!white,
colback=white!10!white,
before title=\raggedright,
title={\textbf{Lecture on Newton’s Second Law of Motion}},
fonttitle=\bfseries,
breakable,
pad at break=1mm
]

\section*{} 

\subsection*{\textcolor{violet}{\textbf{Introduction (5 minutes)}}}
\textbf{Teacher:} \textit{"Good morning, everyone! Today, we’re diving into one of the most fundamental laws of physics—Newton’s Second Law of Motion. This law connects force, mass, and acceleration in a way that explains almost every movement you see around you. By the end of this class, you’ll not only understand the equation but also see how it applies to real-life situations—from pushing a shopping cart to launching a rocket!"}
\textbf{Visual Aid:} Show a short clip of a car accelerating, a ball being kicked, and a rocket taking off.
\textbf{Auditory Cue:} \textit{"Imagine you’re pushing a heavy box versus a light one. Which one moves faster when you push with the same force? That’s what we’re exploring today!"}

\subsection*{\textcolor{violet}{\textbf{Core Explanation (15 minutes)}}}
\subsubsection*{\textcolor{violet}{1. The Law in Words and Equation}}
\textbf{Teacher:} \textit{"Newton’s Second Law states that the acceleration of an object is directly proportional to the net force acting on it and inversely proportional to its mass. Mathematically, it’s written as:"}
\[
F_{\text{net}} = m \cdot a
\]
\textit{Where:}
- $F_{\text{net}}$ = Net force (in Newtons, N)
- $m$ = Mass (in kilograms, kg)
- $a$ = Acceleration (in meters per second squared, m/s²)
\textbf{Intuitive Analogy:} \textit{"Think of force as the ‘push’ you give to a swing. If you push harder (more force), the swing moves faster (more acceleration). But if the swing has a heavy kid (more mass), it won’t speed up as much—even with the same push!"}
\subsubsection*{\textcolor{violet}{2. Breaking Down the Components}}
\textbf{Visual Aid:} Draw a free-body diagram of a car being pushed.
- \textbf{Force (F):} \textit{"The engine’s push or your muscles’ effort."}
- \textbf{Mass (m):} \textit{"How much ‘stuff’ is in the object—more mass means more resistance to motion."}
- \textbf{Acceleration (a):} \textit{"How quickly the speed changes—like a sports car zooming vs. a truck slowly starting."}
\textbf{Real-World Example:}
\textit{"When you kick a soccer ball (light mass), it flies far. But if you kick a bowling ball (heavy mass) with the same force, it barely moves! That’s because acceleration depends on mass."}

\subsection*{\textcolor{violet}{\textbf{Linking Concepts (10 minutes)}}}
\subsubsection*{\textcolor{violet}{1. Net Force vs. Individual Forces}}
\textbf{Teacher:} \textit{"Forces often act in different directions. The ‘net force’ is the overall effect. For example, if you push a box right with 10 N and friction pushes left with 2 N, the net force is 8 N to the right."}
\textbf{Activity:} \textit{"Let’s say a car’s engine provides 5000 N forward, but air resistance is 1000 N backward. What’s the net force?"} (Answer: 4000 N forward.)
\subsubsection*{\textcolor{violet}{2. Units and Calculations}}
\textbf{Teacher:} \textit{"Always check units! If mass is in kg and acceleration in m/s², force must be in Newtons (N)."}
\textbf{Example Problem:}
\textit{"A 2 kg toy car accelerates at 3 m/s². What’s the net force?"}
\[
F = m \cdot a = 2 \, \text{kg} \times 3 \, \text{m/s}^2 = 6 \, \text{N}
\]

\subsection*{\textcolor{violet}{\textbf{Engagement and Practice (10 minutes)}}}
\subsubsection*{\textcolor{violet}{1. Class Discussion}}
\textbf{Teacher:} \textit{"Why does a truck need a bigger engine than a bicycle to reach the same speed?"}
\textit{(Answer: More mass requires more force for the same acceleration.)}
\subsubsection*{\textcolor{violet}{2. Homework Problems}}
1. \textit{"A 50 kg skier accelerates at 2 m/s² downhill. What’s the net force?"}
2. \textit{"If a 1000 kg car accelerates at 0.5 m/s², how much force does the engine provide?"}
3. \textit{"A rocket’s thrust is 10,000 N, and its mass is 2000 kg. What’s its acceleration?"}
\textbf{Advice:} \textit{"Practice these problems tonight! Physics is like a sport—the more you practice, the better you get. Try to relate each problem to real life, like sports or vehicles."}

\subsection*{\textcolor{violet}{\textbf{Summary (5 minutes)}}}
\textbf{Teacher:} \textit{"Let’s recap:}
1. \textit{Newton’s Second Law: $F = m \cdot a$.}
2. \textit{More force $\rightarrow$ more acceleration; more mass $\rightarrow$ less acceleration.}
3. \textit{Net force is the total effect of all forces.}
4. \textit{Always check units and directions!"}
\textit{Next class, we’ll apply this to friction and gravity. Any questions?"}
\textbf{Visual Recap:} Show a slide with the equation, a car accelerating, and a rocket lifting off.

\subsection*{\textcolor{violet}{\textbf{Closing Remarks}}}
\textbf{Teacher:} \textit{"Remember, physics isn’t just about equations—it’s about understanding the world. Observe how objects move around you and ask, ‘What’s the force? What’s the mass?’ That’s how you’ll master this!"}
\textbf{Homework:} Assign 5 problems (mix of calculations and conceptual questions).

\textbf{End of Lecture}
\textbf{Key Takeaways for Students:}
- Understand $F = m \cdot a$ intuitively.
- Relate it to everyday scenarios.
- Practice problems to build confidence.
- Observe real-life applications (e.g., sports, vehicles).
This approach ensures clarity, engagement, and deep understanding!

\end{tcolorbox}

\begin{tcolorbox}[
enhanced jigsaw,
height=\textheight,
size=small,
colframe=blue!50!white,
colback=white!10!white,
before title=\raggedright,
title={\textbf{Lecture on Newton’s Third Law of Motion}},
fonttitle=\bfseries,
breakable,
pad at break=1mm
]

\section*{} 

\subsection*{\textcolor{violet}{\textbf{Introduction (5 minutes)}}}
\textbf{Teacher:} \textit{"Good morning, class! Today, we’re diving into one of the most fascinating laws in physics—Newton’s Third Law of Motion. You’ve probably heard the phrase, ‘For every action, there’s an equal and opposite reaction.’ But what does that really mean? Let’s break it down in a way that makes sense and sticks with you!"}
\textbf{Visual Aid:} Draw a simple diagram of two ice skaters pushing off each other.

\subsection*{\textcolor{violet}{\textbf{Explanation (15 minutes)}}}
\subsubsection*{\textcolor{violet}{1. The Law in Simple Terms}}
- \textbf{Formal Definition:} \textit{"When one object exerts a force on a second object, the second object exerts a force of equal magnitude but in the opposite direction on the first object."}
- \textbf{Intuitive Explanation:} \textit{"Imagine you’re standing on a skateboard and you push against a wall. You move backward, right? That’s because the wall pushes back on you with the same force!"}
\subsubsection*{\textcolor{violet}{2. Key Concepts}}
- \textbf{Action-Reaction Pairs:} Forces always come in pairs—equal in strength, opposite in direction.
- \textbf{They Act on Different Objects:} The forces don’t cancel out because they’re on different bodies.
\textbf{Analogy:} \textit{"Think of a balloon. When you let go, air rushes out (action), and the balloon zooms in the opposite direction (reaction)."}
\textbf{Visual Aid:} Show a rocket launching (thrust pushes exhaust down, exhaust pushes rocket up).

\subsection*{\textcolor{violet}{\textbf{Real-World Examples (10 minutes)}}}
1. \textbf{Walking:} \textit{"When you walk, your foot pushes backward on the ground (action), and the ground pushes you forward (reaction)."}
2. \textbf{Swimming:} \textit{"You push water backward with your arms (action), and the water pushes you forward (reaction)."}
3. \textbf{Car Tires:} \textit{"Tires push backward on the road (action), and the road pushes the car forward (reaction)."}
\textbf{Activity:} \textit{"Pair up and push against each other’s hands. Feel the equal and opposite forces?"}

\subsection*{\textcolor{violet}{\textbf{Common Misconceptions (5 minutes)}}}
- \textit{"If forces are equal, why don’t they cancel out?"}
 - \textbf{Clarification:} \textit{"They act on different objects! Your push on the wall and the wall’s push on you don’t cancel because they’re not on the same body."}

\subsection*{\textcolor{violet}{\textbf{Problem-Solving Practice (10 minutes)}}}
\textbf{Example 1:} \textit{"A book rests on a table. Identify the action-reaction pair."}
- \textbf{Solution:} \textit{"The book pushes down on the table (action), and the table pushes up on the book (reaction)."}
\textbf{Example 2:} \textit{"A bird flies by flapping its wings. Explain using Newton’s Third Law."}
- \textbf{Solution:} \textit{"Wings push air downward (action), and air pushes the bird upward (reaction)."}
\textbf{Homework Assignment:} \textit{"Find 5 real-life examples of Newton’s Third Law and explain them in your own words."}

\subsection*{\textcolor{violet}{\textbf{Summary (5 minutes)}}}
- \textbf{Recap:} \textit{"Newton’s Third Law says forces always come in pairs—equal and opposite, acting on different objects."}
- \textbf{Key Takeaway:} \textit{"Every force has a partner! Look for action-reaction pairs in everyday life."}
\textbf{Final Advice:} \textit{"Practice identifying these pairs in your surroundings. The more you observe, the clearer this law becomes!"}

\subsection*{\textcolor{violet}{\textbf{Closing}}}
\textit{"Great job today! Next class, we’ll tackle some tricky problems together. Until then, keep exploring physics in the world around you!"}
\textbf{Visual Aid:} End with a fun meme or comic strip illustrating the law (e.g., a dog chasing its tail with "action" and "reaction" labels).

\textbf{Note:} Encourage students to ask questions and revisit concepts if needed. Emphasize that understanding > memorization!

\end{tcolorbox}

\begin{tcolorbox}[
enhanced jigsaw,
height=\textheight,
size=small,
colframe=blue!50!white,
colback=white!10!white,
before title=\raggedright,
title={\textbf{Lecture on Plot and Setting}},
fonttitle=\bfseries,
breakable,
pad at break=1mm
]

\section*{} 

\subsection*{\textcolor{violet}{\textbf{Introduction (5 minutes)}}}
\textbf{Teacher:} \textit{"Good morning, class! Today, we’re diving into two of the most fundamental elements of storytelling: \textbf{Plot and Setting}. Think of them as the backbone and the stage of a story. Without them, a story would feel like a puzzle with missing pieces. By the end of this lesson, you’ll not only understand these concepts but also see how they shape the stories you love—whether it’s a book, a movie, or even a video game!"}
\textit{(Pause for a moment, let them settle in.)}
\textit{"Before we begin, let’s do a quick warm-up. Imagine you’re watching a movie. What’s the first thing that grabs your attention? Is it the action? The characters? Or maybe the world they’re in? That’s exactly what we’ll explore today!"}

\subsection*{\textcolor{violet}{\textbf{1. Understanding Plot (15 minutes)}}}
\textbf{Definition:} \textit{"Plot is the sequence of events that make up a story. It’s like the roadmap of a journey—where the story begins, the twists and turns it takes, and how it finally reaches its destination."}
\textbf{Visual Analogy:} \textit{"Imagine you’re playing a video game. The plot is like the main questline—it guides you from the start (the introduction) to the end (the climax and resolution). Along the way, there are side quests (subplots) that add depth to the experience."}
\textbf{Key Components of Plot:}
1. \textbf{Exposition} – \textit{"The setup. Like the first few minutes of a movie where you meet the characters and understand their world."}
2. \textbf{Rising Action} – \textit{"The buildup. Think of it like climbing a rollercoaster—each event raises the stakes."}
3. \textbf{Climax} – \textit{"The peak! The most intense moment, like the final battle in an action movie."}
4. \textbf{Falling Action} – \textit{"The aftermath. What happens right after the big moment?"}
5. \textbf{Resolution} – \textit{"The conclusion. How does everything wrap up?"}
\textbf{Real-World Example:} \textit{"Let’s take \textit{Harry Potter and the Sorcerer’s Stone}.
- \textbf{Exposition}: Harry lives with the Dursleys, unaware he’s a wizard.
- \textbf{Rising Action}: He goes to Hogwarts, learns magic, and discovers the Sorcerer’s Stone.
- \textbf{Climax}: The final confrontation with Voldemort.
- \textbf{Falling Action}: Harry recovers in the hospital wing.
- \textbf{Resolution}: The school year ends, and Harry returns to the Dursleys—but now, everything is different."}
\textbf{Activity:} \textit{"Quick! Think of your favorite movie or book. Can you identify these five stages in it? Discuss with your neighbor for 2 minutes."}

\subsection*{\textcolor{violet}{\textbf{2. Understanding Setting (15 minutes)}}}
\textbf{Definition:} \textit{"Setting is the time and place where a story unfolds. It’s not just a backdrop—it shapes the mood, the characters, and even the plot itself."}
\textbf{Visual Analogy:} \textit{"Imagine setting as the ‘skin’ of a story. A horror story set in a haunted mansion feels very different from one set in a sunny beach town, right? The setting influences how we feel about the story."}
\textbf{Key Aspects of Setting:}
1. \textbf{Time} – \textit{"Is it the past, present, or future? A medieval kingdom or a dystopian future?"}
2. \textbf{Place} – \textit{"A bustling city, a quiet village, or even outer space?"}
3. \textbf{Social Environment} – \textit{"The culture, rules, and norms of the world. For example, \textit{The Hunger Games} has a brutal, oppressive society."}
\textbf{Real-World Example:} \textit{"Let’s compare two stories:
- \textit{The Lion King} (African savanna, vibrant and wild) vs. \textit{The Road} (post-apocalyptic wasteland, bleak and hopeless).
The setting changes how we experience the story!"}
\textbf{Activity:} \textit{"Close your eyes and imagine a story set in a snowy mountain village. What details come to mind? Share with the class."}

\subsection*{\textcolor{violet}{\textbf{3. How Plot and Setting Work Together (10 minutes)}}}
\textit{"Plot and setting are like best friends—they influence each other. A story’s setting can create obstacles for the plot (e.g., a storm delaying a journey) or enhance the mood (e.g., a dark forest making a chase scene scarier)."}
\textbf{Example:} \textit{"In \textit{The Hobbit}, the setting of the Misty Mountains isn’t just a place—it’s a challenge for Bilbo and the dwarves, full of goblins and danger. The plot (their journey) is shaped by the setting."}
\textbf{Discussion Question:} \textit{"How would \textit{Cinderella} change if it were set in modern-day New York instead of a medieval kingdom?"}

\subsection*{\textcolor{violet}{\textbf{Summary (5 minutes)}}}
\textit{"Today, we learned:
1. \textbf{Plot} is the sequence of events (exposition, rising action, climax, etc.).
2. \textbf{Setting} is the time and place, which affects the story’s mood and events.
3. They work together—setting can drive the plot, and the plot can reveal the setting."}
\textbf{Homework:}
1. \textit{Pick a short story or movie and map out its plot structure.}
2. \textit{Rewrite a fairy tale with a completely different setting (e.g., \textit{Snow White} in a cyberpunk city).}
\textbf{Final Advice:} \textit{"Stories are everywhere—books, games, even your own life! The more you practice analyzing plot and setting, the better you’ll understand and create stories. Keep reading, keep imagining, and most importantly, keep practicing!"}

\subsection*{\textcolor{violet}{\textbf{End of Lecture}}}
\textit{"Next class, we’ll dive into character development. Until then, happy storytelling!"}
\textit{(Smile, encourage questions, and dismiss the class.)}

\textbf{Key Takeaways for Teaching Style:}
- \textbf{Slow pace}, \textbf{intuitive analogies}, \textbf{visual examples}, and \textbf{real-world connections}.
- \textbf{Heavy focus on lecture clarity} with \textbf{engaging discussions} and \textbf{homework for practice}.
- \textbf{Encouraged individual practice} beyond class.

\end{tcolorbox}

\begin{tcolorbox}[
enhanced jigsaw,
height=\textheight,
size=small,
colframe=blue!50!white,
colback=white!10!white,
before title=\raggedright,
title={\textbf{Lecture: Structure of Prokaryotic and Eukaryotic Cells}},
fonttitle=\bfseries,
breakable,
pad at break=1mm
]

\section*{} 

\subsection*{\textcolor{violet}{\textbf{Introduction (5 minutes)}}}
\textit{"Good morning, everyone! Today, we’re diving into the fascinating world of cells—the building blocks of life. Imagine cells as tiny cities, each with its own structures and functions. Some cities (prokaryotic cells) are like small, efficient villages, while others (eukaryotic cells) are like bustling metropolises with specialized districts. By the end of this lesson, you’ll be able to tell them apart, understand their key features, and even visualize how they work!"}

\subsection*{\textcolor{violet}{\textbf{1. Prokaryotic Cells: The Simple but Mighty Villages}}}
\textit{(Visual: Draw a simple prokaryotic cell on the board—no nucleus, circular DNA, ribosomes, cell wall, flagella.)}
\textbf{Key Features:}
- \textbf{No nucleus}: DNA floats freely in the cytoplasm (like a village with no city hall—rules are everywhere!).
- \textbf{Circular DNA}: A single loop of genetic material (imagine a rubber band holding all the village’s laws).
- \textbf{Ribosomes}: Tiny protein factories (like bakeries making bread for the village).
- \textbf{Cell wall}: A rigid outer layer (like a protective fence around the village).
- \textbf{Flagella/pili}: Tail-like structures for movement or attachment (like a village’s roads or bridges).
\textbf{Real-World Example:}
- \textit{E. coli} (a bacterium in your gut) is a prokaryote. It’s simple but survives in harsh conditions—just like a resilient village!
\textbf{Analogy:}
\textit{"Think of a prokaryotic cell as a tiny, self-sufficient cabin in the woods. It has everything it needs in one room—no separate bedrooms (nucleus) or fancy kitchens (organelles)."}

\subsection*{\textcolor{violet}{\textbf{2. Eukaryotic Cells: The Complex Metropolises}}}
\textit{(Visual: Draw a eukaryotic cell—nucleus, mitochondria, ER, Golgi, etc.)}
\textbf{Key Features:}
- \textbf{Nucleus}: The "brain" of the cell, storing DNA (like a city hall with a secure vault for laws).
- \textbf{Membrane-bound organelles}: Specialized compartments (like factories, power plants, and post offices).
 - \textbf{Mitochondria}: Powerhouses (like a city’s electrical grid).
 - \textbf{Endoplasmic Reticulum (ER)}: Protein and lipid factories (like assembly lines).
 - \textbf{Golgi apparatus}: Packaging and shipping center (like a post office).
- \textbf{Cytoskeleton}: Structural support (like a city’s roads and bridges).
\textbf{Real-World Example:}
- Your skin cells, liver cells, and even plant cells are eukaryotic. They’re complex but highly efficient!
\textbf{Analogy:}
\textit{"A eukaryotic cell is like New York City—organized, with specialized zones (organelles) for different jobs. The nucleus is City Hall, mitochondria are power plants, and the Golgi is FedEx!"}

\subsection*{\textcolor{violet}{\textbf{3. Comparing Prokaryotes and Eukaryotes}}}
\textit{(Visual: Side-by-side comparison table on the board.)} \\
\begin{tabular}{|l|l|l|}
\hline
\textbf{Feature} & \textbf{Prokaryotic Cell} & \textbf{Eukaryotic Cell} \\
\hline
\textbf{Nucleus} & No nucleus (DNA free) & Nucleus (DNA enclosed) \\
\hline
\textbf{Size} & Small (1–10 $\mu$m) & Large (10–100 $\mu$m) \\
\hline
\textbf{Organelles} & Few or none & Many (mitochondria, ER, etc.) \\
\hline
\textbf{Example} & Bacteria, Archaea & Plants, Animals, Fungi \\
\hline
\end{tabular} \\
\textbf{Activity:}
\textit{"Quick quiz! If I say ‘no nucleus,’ is it prokaryotic or eukaryotic?"} \textit{(Pause for answers.)} \textit{"Correct—prokaryotic! And if I say ‘mitochondria’?"} \textit{(Pause.)} \textit{"Eukaryotic!"}

\subsection*{\textcolor{violet}{\textbf{4. Why Does This Matter?}}}
- \textbf{Medical relevance}: Antibiotics target prokaryotic cells (bacteria) but not your eukaryotic cells.
- \textbf{Evolution}: Eukaryotes evolved from prokaryotes—like villages growing into cities!

\subsection*{\textcolor{violet}{\textbf{Summary (5 minutes)}}}
\textit{"Let’s recap:
1. Prokaryotes = simple, no nucleus (like a village).
2. Eukaryotes = complex, with a nucleus and organelles (like a city).
3. Both are essential—prokaryotes keep ecosystems running, while eukaryotes make up all multicellular life!"}

\subsection*{\textcolor{violet}{\textbf{Homework \& Practice}}}
1. \textbf{Draw and label} a prokaryotic and eukaryotic cell.
2. \textbf{Compare and contrast} the two in a short paragraph.
3. \textbf{Research}: Find one disease caused by prokaryotes (e.g., strep throat) and one by eukaryotic cells (e.g., malaria).
\textbf{Advice:}
\textit{"Practice drawing these cells—visualizing helps! And remember, the more you review, the easier it gets. See you next class!"}

\textbf{End with a smile and open the floor for questions.}

\end{tcolorbox}

\begin{tcolorbox}[
enhanced jigsaw,
height=\textheight,
size=small,
colframe=blue!50!white,
colback=white!10!white,
before title=\raggedright,
title={\textbf{Lecture Plan: Themes in Shakespearean Plays}},
fonttitle=\bfseries,
breakable,
pad at break=1mm
]

\section*{} 

\subsection*{\textcolor{violet}{\textbf{Introduction (5 minutes)}}}
\textbf{Objective:} Set the stage for understanding themes in Shakespeare’s works.
\textbf{Approach:}
- \textit{Visual:} Display a collage of famous Shakespearean play posters (\textit{Romeo and Juliet, Macbeth, Hamlet, A Midsummer Night’s Dream}).
- \textit{Auditory:} Ask students, \textit{"If you had to describe these plays in one word, what would it be?"} (E.g., love, betrayal, revenge, magic).
- \textbf{Analogy:} \textit{"Think of Shakespeare’s plays like a buffet. Each dish (play) has its own flavor (theme), but some ingredients (ideas) keep appearing—like love, power, or fate. Today, we’ll taste-test these themes!"}

\subsection*{\textcolor{violet}{\textbf{Core Lecture (30 minutes)}}}
\textbf{1. Defining Themes}
- \textbf{Technical:} \textit{"A theme is a universal idea explored in a text. It’s not the plot (what happens) but the deeper message (why it matters)."}
- \textbf{Intuitive:} \textit{"If a play is a tree, the plot is the trunk, and themes are the roots—hidden but holding everything up."}
- \textbf{Example:} \textit{Macbeth} isn’t just about a king’s murder; it’s about \textit{ambition} and \textit{guilt}.
\textbf{2. Key Themes in Shakespeare}
\subsubsection*{\textcolor{violet}{A. Love and Conflict}}
- \textit{Visual:} Show a Venn diagram of \textit{Romeo and Juliet} (love vs. family feud).
- \textbf{Real-World Example:} \textit{"Imagine two best friends whose parents hate each other. Their love is pure, but the world around them is toxic—just like Romeo and Juliet."}
\subsubsection*{\textcolor{violet}{B. Power and Corruption}}
- \textit{Auditory:} \textit{"Picture a game of thrones (literally!). In \textit{Macbeth}, the crown is the prize, but winning it destroys the winner."}
- \textbf{Example Problem:} \textit{"If you were Macbeth, would you kill Duncan for power? Why or why not?"} (Discuss in pairs.)
\subsubsection*{\textcolor{violet}{C. Appearance vs. Reality}}
- \textbf{Analogy:} \textit{"Like Instagram filters, characters in \textit{Hamlet} hide their true selves. Claudius smiles but is a murderer; Hamlet pretends madness but is plotting revenge."}
\subsubsection*{\textcolor{violet}{D. Fate and Free Will}}
- \textit{Visual:} Draw a flowchart: \textit{"Romeo’s ‘star-crossed’ fate vs. his choices (e.g., crashing the Capulet party)."}
\textbf{3. Linking Themes Across Plays}
- \textbf{Activity:} \textit{"Match the theme to the play!"} (Hand out slips with themes/plays to pair up.)
- \textbf{Summary:} \textit{"Shakespeare’s themes are like threads in a tapestry—each play weaves them differently, but the patterns connect."}

\subsection*{\textcolor{violet}{\textbf{Engagement \& Homework (10 minutes)}}}
\textbf{1. In-Class Discussion:}
- \textit{"Which theme do you think is most relevant today? Why?"} (Call on 3 students.)
\textbf{2. Homework:}
- \textbf{Example Problems:}
 - \textit{Write a paragraph comparing a Shakespearean theme to a modern movie (e.g., \textit{The Lion King} and \textit{Hamlet}).}
 - \textit{Create a meme or comic strip showing a theme from \textit{A Midsummer Night’s Dream}.}
\textbf{3. Advice for Practice:}
- \textit{"Read play summaries online, then ask: ‘What’s the lesson here?’ Practice spotting themes in songs or TV shows too!"}

\subsection*{\textcolor{violet}{\textbf{Summary (5 minutes)}}}
\textbf{Recap:}
- Themes are the "big ideas" in plays.
- Shakespeare’s favorites: love, power, deception, fate.
- \textbf{Final Analogy:} \textit{"Themes are like spices—you’ll taste them in every dish (play) Shakespeare cooks up!"}
\textbf{Exit Ticket:} \textit{"Name one theme and one play where it appears."}

\subsection*{\textcolor{violet}{\textbf{Teaching Notes:}}}
- \textbf{Pace:} Slow, with pauses for questions.
- \textbf{Tone:} Conversational but structured.
- \textbf{Visuals:} Use slides, diagrams, and props (e.g., a toy crown for \textit{Macbeth}).
- \textbf{Follow-Up:} Next class, discuss homework examples to reinforce understanding.
\textbf{Goal:} By the end, students should see themes as the "DNA" of Shakespeare’s works—essential, recurring, and deeply human.

\end{tcolorbox}

\begin{tcolorbox}[
enhanced jigsaw,
height=\textheight,
size=small,
colframe=blue!50!white,
colback=white!10!white,
before title=\raggedright,
title={\textbf{Lecture Plan: Trigonometric Identities}},
fonttitle=\bfseries,
breakable,
pad at break=1mm
]

\section*{} 

\subsection*{\textcolor{violet}{\textbf{Introduction (10 minutes)}}}
\textbf{Objective:} Build intuition for trigonometric identities using real-world analogies.
\textbf{Approach:}
\textit{Visual:} Draw a unit circle on the board. Label angles ($\theta$) and coordinates ($\cos\theta$, $\sin\theta$).
\textit{Intuitive Analogy:} "Imagine a Ferris wheel. At any point, your height above the ground is like the sine of the angle, and your horizontal distance from the center is like the cosine. No matter where you are on the wheel, the relationship between your height and distance follows a pattern—just like identities!"
\textit{Auditory:} Use a conversational tone: "Why do we need identities? Because they’re like shortcuts in math—just like how you’d use a recipe to bake a cake instead of figuring it out from scratch every time."
\textbf{Key Points:}
Identities are equations true for all angles.
They simplify complex trigonometric expressions.

\subsection*{\textcolor{violet}{\textbf{Core Concepts (20 minutes)}}}
\textbf{Objective:} Derive and explain fundamental identities with visuals and examples.

\subsubsection*{\textcolor{violet}{\textbf{A. Pythagorean Identity}}}
\textit{Visual:} Show a right triangle inscribed in the unit circle.
Hypotenuse = 1 (radius).
Legs = $\cos\theta$ and $\sin\theta$.
By Pythagoras: $\sin^2\theta + \cos^2\theta = 1$.
\textit{Real-World Example:} "A ladder leaning against a wall: the height ($\sin\theta$) and base ($\cos\theta$) change, but the ladder’s length (1) stays the same."

\subsubsection*{\textcolor{violet}{\textbf{B. Reciprocal Identities}}}
\textit{Analogy:} "Like flipping a fraction—cosecant is just 1/sine, like how ‘inverse’ means ‘opposite’ in everyday language."
\textit{Visual:} Write identities on the board:
$\csc\theta = \frac{1}{\sin\theta}$, $\sec\theta = \frac{1}{\cos\theta}$, $\cot\theta = \frac{1}{\tan\theta}$.

\subsubsection*{\textcolor{violet}{\textbf{C. Quotient Identity}}}
\textit{Derivation:} $\tan\theta = \frac{\sin\theta}{\cos\theta}$.
\textit{Example:} "If you’re climbing a slope, $\tan\theta$ is the ratio of how high you go ($\sin\theta$) to how far you move forward ($\cos\theta$)."

\subsection*{\textcolor{violet}{\textbf{Linking Concepts (10 minutes)}}}
\textbf{Objective:} Show how identities connect to each other.

\textit{Activity:} "Let’s take $\sin^2\theta + \cos^2\theta = 1$. Divide both sides by $\sin^2\theta$. What do you get?"
Derive $1 + \cot^2\theta = \csc^2\theta$.
\textit{Visual:} Draw a flowchart linking Pythagorean to other identities.

\subsection*{\textcolor{violet}{\textbf{Practice \& Engagement (15 minutes)}}}
\textbf{Objective:} Reinforce understanding through examples.

\textbf{1. Example Problems (Solve Together):}
Simplify $\frac{\sin\theta}{\tan\theta}$.
Prove $\sec\theta - \cos\theta = \sin\theta \tan\theta$.
\textbf{2. Homework Assignment:}
5 problems (e.g., verify identities, simplify expressions).
\textit{Advice for Practice:} "Practice is like training for a sport—the more you do, the stronger your skills get!"

\subsection*{\textcolor{violet}{\textbf{Summary \& Advice (5 minutes)}}}
\textbf{Key Takeaways:}
Identities are universal truths in trigonometry.
Use the unit circle and right triangles to visualize them.
Practice is essential—work on problems daily!
\textit{Final Note:} "Trigonometry is like a puzzle. The more pieces (identities) you know, the easier it is to solve. Keep practicing!"

\subsection*{\textcolor{violet}{\textbf{Teaching Style Recap:}}}
\textbf{Pace:} Slow, with pauses for questions.
\textbf{Tone:} Friendly but precise.
\textbf{Visuals:} Heavy use of diagrams and real-world parallels.
\textbf{Engagement:} Focus on guided examples and homework.

\subsection*{\textcolor{violet}{\textbf{Follow-Up:}}}
Next class will cover angle addition formulas—bring your questions!
\textbf{Goal:} By the end, students should have a solid grasp of fundamental trigonometric identities and their applications.

\end{tcolorbox}

\begin{tcolorbox}[
enhanced jigsaw,
height=\textheight,
size=small,
colframe=blue!50!white,
colback=white!10!white,
before title=\raggedright,
title={\textbf{Lecture Plan: Trigonometric Ratios}},
fonttitle=\bfseries,
pad at break=1mm
]
{
\fontsize{11}{12}\selectfont
\section*{} 

\subsection*{\textcolor{violet}{\textbf{Objective:}}}
By the end of this lesson, students will understand the fundamental trigonometric ratios ($\sin\theta$, $\cos\theta$, $\tan\theta$) and their applications in real-world scenarios. They will be able to visualize these ratios in right-angled triangles and solve problems involving them.

\subsection*{\textcolor{violet}{\textbf{1. Introduction (5 minutes)}}}
\textbf{Teacher:}
"Imagine you're standing at the base of a tall building, looking up at the top. How could you measure its height without climbing it? Or suppose you're a ship captain navigating the ocean—how do you calculate distances using angles? Today, we’ll learn about trigonometric ratios, which are the secret tools mathematicians and engineers use to solve such problems!"
\textbf{Visual Aid:}
Draw a right-angled triangle on the board with sides labeled (opposite, adjacent, hypotenuse).

\subsection*{\textcolor{violet}{\textbf{2. Core Concept: Trigonometric Ratios (15 minutes)}}}
\textbf{Definition:}
In a right-angled triangle, the three primary trigonometric ratios are:
1. $\sin\theta = \frac{\text{Opposite}}{\text{Hypotenuse}}$
2. $\cos\theta = \frac{\text{Adjacent}}{\text{Hypotenuse}}$
3. $\tan\theta = \frac{\text{Opposite}}{\text{Adjacent}}$
\textbf{Analogy:}
"Think of a ladder leaning against a wall. The angle it makes with the ground is $\theta$. The height it reaches (opposite side) depends on how steep the angle is. If you know the angle and the length of the ladder (hypotenuse), you can find the height using sine!"
\textbf{Visualization:}
Draw a ladder against a wall (right triangle).
Label sides: hypotenuse (ladder), opposite (height), adjacent (distance from wall).
\textbf{Real-World Example:}
"Astronomers use trigonometry to measure the distance to stars. If they know the angle of elevation and the baseline distance, they can calculate how far a star is!"

\subsection*{\textcolor{violet}{\textbf{3. Step-by-Step Explanation (20 minutes)}}}
\textbf{Example Problem:}
"A flagpole casts a 10-meter shadow when the sun is at a 30$^\circ$ angle. How tall is the flagpole?"
\textbf{Solution:}
1. Identify the sides:
Opposite = height of the flagpole (unknown).
Adjacent = shadow length (10 m).
Angle $\theta$ = 30$^\circ$.
2. Use $\tan\theta = \frac{\text{Opposite}}{\text{Adjacent}}$.
$\tan(30^\circ) = \frac{\text{height}}{10}$.
height = $10 \times \tan(30^\circ)$.
height $\approx 10 \times 0.577 \approx 5.77$ meters.
\textbf{Interactive Check:}
"If the angle were steeper (say 45$^\circ$), would the flagpole be taller or shorter? Why?"

\subsection*{\textcolor{violet}{\textbf{4. Common Mistakes \& Clarifications (10 minutes)}}}
\textbf{Mistake:} Confusing opposite and adjacent sides.
\textbf{Fix:} "Always draw the triangle and label sides relative to the angle $\theta$!"
\textbf{Mistake:} Forgetting to use the correct ratio.
\textbf{Fix:} "Remember SOH-CAH-TOA!"
SOH: $\sin\theta = \frac{\text{Opposite}}{\text{Hypotenuse}}$
CAH: $\cos\theta = \frac{\text{Adjacent}}{\text{Hypotenuse}}$
TOA: $\tan\theta = \frac{\text{Opposite}}{\text{Adjacent}}$

\subsection*{\textcolor{violet}{\textbf{5. Summary (5 minutes)}}}
\textbf{Key Takeaways:}
Trigonometric ratios relate angles to sides in right triangles.
\textbf{SOH-CAH-TOA} is your best friend for remembering ratios.
Real-world applications: architecture, navigation, astronomy.
\textbf{Homework Assignment:}
Solve 5 problems (e.g., finding heights, distances, angles).
Practice drawing triangles and labeling sides.
\textbf{Advice:}
"Trigonometry is like a muscle—the more you practice, the stronger your understanding becomes. Try solving problems daily!"

\subsection*{\textcolor{violet}{\textbf{6. Closing Remarks}}}
"Next class, we’ll explore how these ratios work for non-right triangles. Until then, keep practicing!"
\textbf{Visual Recap:}
Show a slide with:
Right triangle labeled with sides and angle.
SOH-CAH-TOA mnemonic.
Example problem solved step-by-step.

\subsection*{\textcolor{violet}{\textbf{Teaching Style Notes:}}}
\textbf{Pace:} Slow, with pauses for questions.
\textbf{Tone:} Friendly, conversational.
\textbf{Engagement:} Heavy focus on lecture + examples.
\textbf{Visuals:} Diagrams, real-world images (e.g., ladders, flagpoles).
\textbf{Homework:} Emphasize individual practice.
}
\end{tcolorbox}

\begin{tcolorbox}[
enhanced jigsaw,
height=\textheight,
size=small,
colframe=blue!50!white,
colback=white!10!white,
before title=\raggedright,
title={\textbf{Lecture on Valency and Lewis Structures}},
fonttitle=\bfseries,
pad at break=1mm
]
{
\fontsize{11}{12}\selectfont
\section*{} 

\subsection*{\textcolor{violet}{\textbf{Introduction (5 minutes)}}}
\textbf{Teacher:} "Good morning, class! Today, we’re going to explore two fundamental concepts in chemistry: \textbf{Valency} and \textbf{Lewis Structures}. These ideas help us understand how atoms bond to form molecules—the building blocks of everything around us! By the end of this lesson, you’ll be able to predict how atoms connect and even draw simple molecular structures. Let’s dive in!"
\textit{Visual:} Show a slide with a simple molecule like $\text{H}_2\text{O}$ and ask students what they notice about the arrangement of atoms.

\subsection*{\textcolor{violet}{\textbf{1. Understanding Valency (15 minutes)}}}
\textbf{Definition:} "Valency is the combining power of an atom—how many bonds it can form with other atoms. Think of it like the number of ‘hands’ an atom has to hold onto other atoms!"
\textbf{Analogy:} "Imagine atoms are like LEGO blocks. Some blocks have one connector (like hydrogen), while others have four (like carbon). Valency tells us how many connectors an atom has!"
\textbf{Real-World Example:}
\textbf{Hydrogen (H):} Valency = 1 (forms 1 bond, e.g., $\text{H}_2$ or $\text{H}_2\text{O}$).
\textbf{Oxygen (O):} Valency = 2 (forms 2 bonds, e.g., $\text{H}_2\text{O}$ or $\text{CO}_2$).
\textbf{Carbon (C):} Valency = 4 (forms 4 bonds, e.g., $\text{CH}_4$ or $\text{CO}_2$).
\textit{Visual:} Show a table of common elements with their valencies. Draw stick-and-ball models of $\text{H}_2\text{O}$ and $\text{CH}_4$ on the board.
\textbf{Activity:} "Quick check! If nitrogen (N) has a valency of 3, how many hydrogen atoms can it bond with to form ammonia ($\text{NH}_3$)?" \textit{Answer: 3!}

\subsection*{\textcolor{violet}{\textbf{2. Lewis Structures: Drawing Molecular Bonds (20 minutes)}}}
\textbf{Definition:} "Lewis Structures are diagrams that show how atoms bond by sharing electrons. They use dots (for lone pairs) and lines (for bonds)."
\textbf{Steps to Draw a Lewis Structure:}
1. Count valence electrons (use the periodic table).
2. Arrange atoms (central atom is usually the least electronegative).
3. Place electrons (fill octets, except hydrogen which needs 2).
\textbf{Example: Water ($\text{H}_2\text{O}$)}
1. \textbf{Oxygen (O):} 6 valence electrons.
2. \textbf{Hydrogen (H):} 1 valence electron each (total = 2).
3. \textbf{Total electrons:} 6 + 2 = 8.
4. \textbf{Draw:} O in the center, two H atoms attached, and lone pairs to complete the octet.
\textit{Visual:} Draw $\text{H}_2\text{O}$ step-by-step on the board. Show how oxygen shares electrons with hydrogen.
\textbf{Analogy:} "Think of Lewis Structures like a dance floor. Atoms are dancers, and electrons are the music keeping them together. Each atom wants to ‘dance’ (bond) until it’s happy (full octet)!"
\textbf{Real-World Example:}
\textbf{Carbon Dioxide ($\text{CO}_2$):} Carbon (4 valence) bonds with two oxygens (6 each).
\textbf{Methane ($\text{CH}_4$):} Carbon (4 valence) bonds with four hydrogens (1 each).
\textit{Visual:} Show $\text{CO}_2$ and $\text{CH}_4$ Lewis structures. Highlight double bonds in $\text{CO}_2$.
\textbf{Activity:} "Let’s try one together! Draw the Lewis structure for ammonia ($\text{NH}_3$)." \textit{Guide students through the steps.}

\subsection*{\textcolor{violet}{\textbf{3. Common Mistakes and Tips (5 minutes)}}}
\textbf{Mistake:} Forgetting lone pairs (e.g., oxygen in $\text{H}_2\text{O}$ needs 2 lone pairs).
\textbf{Tip:} Always count valence electrons first!
\textbf{Mistake:} Drawing incorrect central atoms (e.g., H is never central).
\textit{Visual:} Show incorrect vs. correct Lewis structures for $\text{NH}_3$.

\subsection*{\textcolor{violet}{\textbf{4. Summary and Homework (5 minutes)}}}
\textbf{Summary:}
Valency = number of bonds an atom can form.
Lewis Structures show bonding and lone pairs.
Key steps: Count electrons, arrange atoms, fill octets.
\textbf{Homework:}
1. Draw Lewis structures for: $\text{H}_2$, $\text{O}_2$, $\text{N}_2$, $\text{HCl}$, $\text{CCl}_4$.
2. Predict the valency of sulfur (S) and phosphorus (P) using the periodic table.
\textbf{Advice:} "Practice makes perfect! Try drawing these structures at home, and don’t hesitate to ask questions in our next class."

\subsection*{\textcolor{violet}{\textbf{Closing (2 minutes)}}}
\textbf{Teacher:} "Great job today! Remember, chemistry is like solving puzzles—once you see the pattern, it becomes fun. Next time, we’ll explore how these bonds create different types of molecules. See you then!"
\textit{Visual:} End with a slide of a complex molecule like glucose, hinting at future topics.

\subsection*{\textcolor{violet}{\textbf{Key Teaching Notes:}}}
\textbf{Pace:} Slow, with pauses for questions.
\textbf{Engagement:} Heavy focus on visuals and analogies.
\textbf{Practice:} Emphasize homework and individual study.
}
\end{tcolorbox}
\end{document}